\def\year{2022}\relax
\documentclass[letterpaper]{article} %
\usepackage{aaai22}  %
\usepackage{times}  %
\usepackage{helvet}  %
\usepackage{courier}  %
\usepackage[hyphens]{url}  %
\usepackage{graphicx} %
\usepackage{natbib}  %
\usepackage{caption} %
\DeclareCaptionStyle{ruled}{labelfont=normalfont,labelsep=colon,strut=off} %
\usepackage{todonotes} %
\usepackage{bm}
\usepackage{amsfonts,amsmath,amsthm}
\usepackage{cleveref}
\newcommand{\triple}[3]{#2(#1, #3)}  %
\newcommand{\temptriple}[4]{#2(#1, #3 | #4)} %

\usepackage{enumitem} 
\usepackage{makecell}
\usepackage{subcaption}
\usepackage{booktabs}
\newtheorem{theorem}{Theorem}[section]

\newtheorem{example}{Example}[section]
\newtheorem{lemma}[theorem]{Lemma}
\title{Temporal Knowledge Graph Completion using Box Embeddings}
\author{
        Johannes Messner, Ralph Abboud, {\.I}smail {\.I}lkan Ceylan\\
}
\affiliations{
    Department of Computer Science\\
    University of Oxford, UK\\
    firstname.lastname@cs.ox.ac.uk
}

\begin{document}

\maketitle

\begin{abstract}
Knowledge graph completion is the task of inferring missing facts based on existing data in a knowledge graph. Temporal knowledge graph completion~(TKGC) is an extension of this task to temporal knowledge graphs, where each fact is additionally associated with a \emph{time stamp}. %
Current approaches for TKGC primarily build on existing embedding models which are developed for (static) knowledge graph completion, and extend these models to incorporate time, where the idea is to learn latent representations for entities, relations, and time stamps and then use the learned representations to predict missing facts at various time steps.
In this paper, we propose BoxTE, a box embedding model for TKGC, building on the static knowledge graph embedding model BoxE. We show that BoxTE is fully expressive, and possesses strong inductive capacity in the temporal setting. We then empirically evaluate our model and show that it achieves state-of-the-art results on several TKGC benchmarks.
\end{abstract}

\section{Introduction}
{Knowledge graphs}~(KGs) are sets of (binary) facts of the form $\triple{h}{r}{t}$, where $h$ and $t$ represent \emph{entities}, and $r$ represents a \emph{relation} that holds between these entities.
KGs play an increasingly prominent role in \emph{representing}, \emph{storing}, and \emph{processing} information. KGs such as YAGO \cite{MahdisoltaniBS15}, Knowledge Vault \cite{GoogleVault}, Freebase \cite{BollackerCT07} and NELL \cite{MitchellBCM18} store hundreds of millions of facts, and are key drivers for downstream tasks, such as \emph{question answering}~\cite{BordesCW14}, \emph{information retrieval}~\cite{XiongPC17}, and \emph{recommender systems}~\cite{WangZWZLXG18}.

Notably, most KGs are inherently \emph{incomplete}, which negatively affects their use in downstream applications. This has motivated a large body of work for automatically inferring missing facts from a given KG, a task known as
\emph{knowledge graph completion}~(KGC).
One prominent approach for KGC is based on KG {embedding models}, where the idea is to learn embeddings for entities and relations through training over known facts, and subsequently use the learned embeddings to compute plausibility scores for all possible facts~\cite{TransE-NIPS13}. 

Standard KG embedding models, however, operate under the assumption that the input KG is \emph{static}, i.e., the ground truth of a fact is independent of time. This assumption is not always realistic, as, e.g., a person living in a location could move to a different location, invalidating the associated fact in the future. Indeed, many real-world KGs include temporal information for their facts, most commonly in the form of time stamps, indicating \emph{when} the fact holds~\cite{KazemiGJKSFP20}. 
Specifically, a \emph{temporal (binary) fact} is a fact of the form $\temptriple{h}{r}{t}{\tau}$, where $h$ and $t$ represent \emph{entities}, $\tau$ is a \emph{time stamp}, and $r$ represents a \emph{relation} that holds between these entities at the time specified by $\tau$. A \emph{temporal knowledge graph}~(TKG) is then a (finite) set of temporal facts.

The focus of this work is on \emph{temporal knowledge graph completion}~(TKGC) which is the task of inferring missing temporal facts from a TKG. The main challenge in TKGC is to additionally learn embeddings for time stamps, such that embedding models perform scoring jointly based on relation, entity and time stamp embeddings. This perspective has led to the development of several embedding models, building on static KGE models~\cite{TTransE,Lacroix2020Tensor}, or having dedicated neural architectures~\cite{wu2020temp, garcia-duran-etal-2018-learning}, to appropriately represent temporal information. However, no current embedding model, to our knowledge, studies TKGC from the perspective of capturing \emph{temporal inference patterns} %
despite their prevalence in real-world data \cite{FB15k237TC}. 

In this paper, we propose BoxTE, a box embedding model for temporal knowledge graph completion. BoxTE builds on the static KG embedding model BoxE \cite{BoxE-NeurIPS20}, and extends it with dedicated time embeddings, allowing to flexibly represent temporal information. In BoxTE, time embeddings are unique for each time point, but they are specific for every relation, which yields a very flexible and powerful representation.  Our contibutions are manifold. We first show that BoxTE is fully expressive, and has strong inductive capacity, capturing, e.g., a rich class of \emph{rigid} inference patterns, and \emph{cross-time} inference patterns. We are not aware of any other model with such capacity, and our study presents the first thorough analysis of inductive capacity in the context of TKGC. 
Empirically, we conduct a detailed experimental evaluation, and show that BoxTE achieves state-of-the-art performance on several TKGC baselines, even with a limited number of parameters. %
Finally, in a dedicated ablation study, we validate our proposed temporal mechanism by comparing with alternate model variations, inspired by existing literature, and show the strength of our proposal.

\section{Temporal Knowledge Graph Completion}

In what follows, we consider a \emph{relational vocabulary}, which consists of a finite set $\mathbf{E}$ of \emph{entities}, and a finite set $\mathbf{R}$ of \emph{relations}. We additionally consider a finite set of \emph{time stamps} $\mathbf{T}$. We denote \emph{static binary facts} as $\triple{h}{r}{t}$, and \emph{temporal binary facts} as $\temptriple{h}{r}{t}{\tau}$, where $h, t \in \mathbf{E}$ are entities, $r \in \mathbf{R}$ is a binary relation, and $\tau \in \mathbf{T}$ is a time stamp. Temporal facts are also referred as \emph{quadruples}; and, for a temporal fact $\temptriple{h}{r}{t}{\tau}$, $h$ is referred to as the \emph{head entity} and $t$ as the \emph{tail entity}, a convention inherited from static binary facts.

A \emph{knowledge graph}~(KG) $\mathcal{G}$ consists of a finite set of binary facts over ($\mathbf{E}, \mathbf{R}$); or equivalently, a KG is a multi-labeled graph, where the  nodes correspond to entities, and labels correspond to relations. A \emph{temporal knowledge graph}~(TKG) $\mathcal{G}$ consists of a finite set of temporal binary facts over ($\mathbf{E}, \mathbf{R}, \mathbf{T}$). 
Given a TKG $\mathcal{G}$, \emph{temporal knowledge graph completion}~(TKGC) is the task of predicting new, unseen facts over ($\mathbf{E}, \mathbf{R}, \mathbf{T}$) based on existing facts in $\mathcal{G}$. 

Typically, TKGC models define a \emph{scoring function} for temporal facts, and are optimized to score true facts in $\mathcal{G}$ higher than corrupted, \emph{negatively sampled} facts. This negative sampling produces \emph{corrupted facts} from a true fact $\temptriple{h}{r}{t}{\tau}$ from $\mathcal{G}$ by replacing either $h$ or $t$ with a random entity $h' \neq h$ (resp., $t' \neq t$).
Empirically, TKGC models are evaluated by scoring true facts from the TKG test set, and then comparing with scores for all possible corrupted facts not appearing in the training, validation, or test set. Using all scores, standard metrics \cite{TransE-NIPS13} are then computed, and these include mean rank (MR), the average rank of facts against their corrupted counterparts, mean reciprocal rank (MRR), their average inverse rank (i.e., 1/rank), and  Hits@K, the proportion of facts with rank at most K. 

Conceptually, TKGC models can be characterized by their \emph{expressiveness} and \emph{inductive capacity}. More specifically, a model is \emph{fully expressive} if, for any given disjoint sets of \emph{true} and \emph{false} facts, it admits a configuration that accurately classifies all facts. On the other hand, \emph{inductive capacity} describes the \emph{inference patterns} that a model can learn and capture. Example inference patterns include relation symmetry, hierarchy and mutual exclusion. Both expressiveness and inductive capacity are key for TKGC, as the former enables fitting of the input TKG, whereas the latter offers a strong inductive bias, and thus improves model generalization.    

\section{Related Work}
\label{sec:rw}

\noindent\textbf{Knowledge graph embedding models.} Knowledge graph embedding (KGE) models represent entities and relations in a KG using \emph{embeddings}, which in turn are learned from data to compute scores for all possible KG facts. KGE models can broadly be classified into translational, bilinear, and neural models. Translational models, such as TransE \cite{TransE-NIPS13} and RotatE \cite{RotatE-ICLR19}, represent entities as points in a low-dimensional space, relations as translations or rotations in this space, and score binary facts based on the distance between the relation-translated head embedding and the tail embedding. A variation on this approach is spatio-translational models, such as BoxE \cite{BoxE-NeurIPS20}, where fact correctness depends on \emph{absolute} representation position in the embedding space. Bilinear models, such as RESCAL \cite{RESCAL-ICML11}, TuckER \cite{TuckER} and ComplEx \cite{ComplEx} are based on \emph{tensor factorization}: they embed entities as vectors, and relations as matrices or tensors, such that the bilinear product between entity and relation embeddings yields fact plausibility scores. Finally, neural models, such as rGCN \cite{SchlichtkrullKB18} and ConvE \cite{ConvE-AAAI18} use a
neural architecture to perform scoring over KG embeddings. KGE models have widely been investigated in recent years, but these models assume that facts are static, and thus do not incorporate temporal information. 

\noindent\textbf{Temporal knowledge graph embedding models.} Analogously to KGE models, temporal knowledge graph embedding (TKGE) models use embeddings to represent entities, relations, and time stamps in a TKG, and subsequently perform fact scoring. Most TKGE models hence build on existing KGE models. For instance, TTransE \cite{TTransE} extends TransE, and encodes time stamps as translations, analogously to relations, such that these translations additionally move head representations in the embedding space. ChronoR \cite{ChronoR} builds on RotatE, and represents time-relation pairs with rotation and scaling in the embedding space. Concretely,  relation and time stamp representations are concatenated to yield an overall rotation vector applied on entity representations. Furthermore, TComplEx and TNTComplEx \cite{Lacroix2020Tensor} are based on ComplEx, and analogously factorize the input TKGC, which both models represent as a fourth-order tensor. TComplEx applies this factorization directly, whereas TNTComplEx divides this factorization into a temporal and a non-temporal component.

In addition to KGE model extensions, other TKGE models process temporal information initially, before subsequently passing the resulting time-conditioned representation to a static KGE model. For example, TeRo \cite{xu2020tero} represents time as a rotation in complex vector space that applies on entity embeddings, following which the TransE scoring mechanism applies. Moreover, HyTE \cite{HyTE} represents each time stamp as a learnable hyper-plane in $d$-dimensional space, and then projects entity and relation embeddings into this hyper-plane and applies the TransE scoring function on the projections. Additionally, diachronic embeddings \cite{DiachronicEmbedding} have been proposed to map entity and relation embeddings, paired with temporal information, into a KGE model space, thus defining a framework yielding specific models such as DE-TransE and DE-SimplE.

Finally, dedicated neural architectures have been applied to exploit the sequential structure inherent in time, and to leverage the graph structure present in temporal knowledge graphs. More specifically, recurrent neural networks (RNNs) \cite{garcia-duran-etal-2018-learning} have been aplied for TKGC, such that these networks model changes in relations over time, and encode the sequential nature of time. More recently, \emph{TeMP} \cite{wu2020temp} leverages message passing graph neural networks (MPNNs) to learn structure-based entity representations at every time stamp, and then aggregates representations across all time stamps using an encoder, yielding the models TeMP-GRU (gated recurrent unit encoder) and TeMP-GRU (self-attention encoder). Similarly to other models, the final entity representations can subsequently be used with a static KGE model, e.g., TransE.

\section{A Temporal Box Embedding Model}

In this section, we introduce BoxTE, a box embedding model for temporal knowledge graph completion. BoxTE builds on the static BoxE model \cite{BoxE-NeurIPS20}, and extends it with a \emph{temporal representation}, which allows to additionally capture inference patterns \emph{across time} and model certain temporal relational information. We first briefly recall BoxE, and then present BoxTE.

\subsection{The box embedding model BoxE} \label{sec:boxe}

\label{ssec:boxe}
BoxE \cite{BoxE-NeurIPS20} is a spatio-translational knowledge base completion model that can predict facts across relations of arbitrary arity. For our purposes, we only present BoxE restricted to the binary setting.

\paragraph{Representation.} In BoxE, every entity $h \in \mathbf{E}$ is represented by two vectors, a \emph{base position} vector $\bm{e_h} \in \mathbb{R}^d,$ and a \emph{translational bump} vector $\bm{b_h} \in \mathbb{R}^d$ , which translates the entity that co-occurs in a fact with $h$. 
For a binary fact $\triple{h}{r}{t}$, the \emph{final embeddings} of the head entity $h$ and the tail entity $t$ are respectively given as:
\begin{align*}
\bm{e_h^{\triple{h}{r}{t}}} = \bm{e_h} + \bm{b_{t}},  \quad \quad \quad \quad
\bm{e_t^{\triple{h}{r}{t}}} = \bm{e_t} + \bm{b_{h}}.
\end{align*}
BoxE represents every binary relation $r$ with two $d$-dimensional boxes: a head box $\bm{r^{h}}$, and a tail box $\bm{r^{t}}$. Semantically, a fact $\triple{h}{r}{t}$ is considered true if final entity representations for $h$ and $t$ appear in their corresponding boxes:
\begin{align*}
 \bm{e_h^{\triple{h}{r}{t}}} \in \bm{r^h}, \quad\quad \bm{e_t^{\triple{h}{r}{t}}} \in \bm{r^t}.
\end{align*}
\paragraph{Scoring.} BoxE scoring function for a true fact $\triple{h}{r}{t}$ encourages box membership, and is defined as:
\begin{align*}
    \mathsf{score}(\triple{h}{r}{t}) = &\left\Vert \delta(\bm{e_h^{\triple{h}{r}{t}}}, \bm{r^{h}})  \right\Vert_x + 
    &\left\Vert \delta(\bm{e_t^{\triple{h}{r}{t}}}, \bm{r^{t}})  \right\Vert_x,
\end{align*}
where $\delta$ intuitively computes the distance between a point and a box, and $x$ indicates the L-$x$ norm. 

\paragraph{Properties.}
BoxE has several desirable properties. First, it is fully expressive%
. Second, it captures a wide array of inference patterns, and thus has strong inductive capacity. For more details and illustrations, we refer the reader to the original paper \cite{BoxE-NeurIPS20}.

\subsection{The temporal box embedding model BoxTE}

We now introduce BoxTE, a box embedding model for temporal knowledge graph completion. At a high level, BoxTE extends BoxE with \emph{time bumps} to represent time stamps in an input knowledge graph, such that these time bumps additionally translate final entity representations. However, unlike entity bumps, time bumps are \emph{not} standard learnable embeddings, but are \emph{induced} by the \emph{relation} of a given target fact, based on a set of \emph{time stamp embeddings}. 

\paragraph{Representation.} In addition to entity and relation representations, BoxTE defines, (i) for every timestamp $\tau \in \mathbf{T}$, a set of $k$ $d$-dimensional embeddings, represented by a matrix $\mathbf{K}^\tau \in \mathbb{R}^{k \times d}$, and (ii) for every relation $r$, a $k$-dimensional scalar vector $\bm{\alpha^r}$. Then, for every time stamp $\tau \in \mathbf{T}$ and relation $r \in \mathbf{R}$, a corresponding \emph{time bump} is given by:
\begin{align*} 
\bm{\tau^{r}}= \bm{\alpha^r} \mathbf{K^\tau}
\end{align*}
For a temporal fact $\temptriple{h}{r}{t}{\tau}$, the \emph{final entity representations} for $h$ and $t$ are given as:
\begin{align*} 
\bm{e_h^{\temptriple{h}{r}{t}{\tau}}} &=\bm{e_h} + \bm{b_t} + \bm{\tau^{r}}, \quad
\bm{e_t^{\temptriple{h}{r}{t}{\tau}}} &=\bm{e_t} + \bm{b_h} + \bm{\tau^{r}}.
\end{align*}
Scoring is then performed analogously to BoxE. Intuitively, time bumps produce distinct final embeddings at every time stamp. However, within this time stamp, every relation also induces potentially distinct entity representations. In particular, time bumps induce distinct final embeddings for $h$ and $t$ for facts $\temptriple{h}{r}{t}{\tau}$ and $\temptriple{h}{s}{t}{\tau}$, due to the distinct scalars of relations $r$ and $s$, respectively. Therefore, time bumps in BoxTE represent \emph{relation-specific temporal dynamics} by learning appropriate scalars $\bm{\alpha^r}$. %

\begin{figure*}
    \centering
    \includegraphics[width=\textwidth]{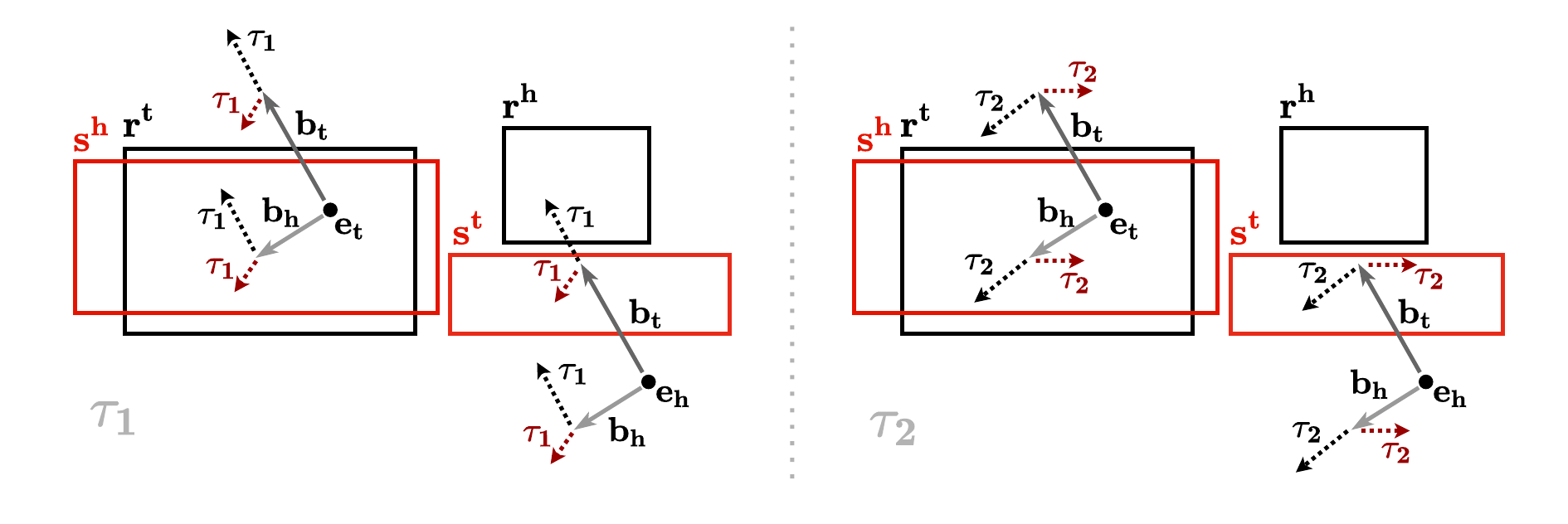}
    \caption{A sample BoxTE configuration for the TKG $\mathcal{G}$ at time stamps $\tau_1$ (left) and $\tau_2$ (right). Time bumps $\bm{\tau_1^r}$, $\bm{\tau_2^r}$, $\bm{\tau_1^s}$ and $\bm{\tau_2^s}$ are computed as $\bm{\tau_1^r}=\bm{\alpha^r}\mathbf{K^{\tau_1}}$, $\bm{\tau_2^r}=\bm{\alpha^r}\mathbf{K^{\tau_2}}$, $\bm{\tau_1^s}=\bm{\alpha^s}\mathbf{K^{\tau_1}}$, and $\bm{\tau_2^s}=\bm{\alpha^s}\mathbf{K^{\tau_2}}$, respectively, and represented by dotted arrows. Relation $r$ and corresponding time bumps $\bm{\tau_1^r}$, $\bm{\tau_2^r}$ are color coded with black, whereas the relation $s$, and the corresponding time bumps $\bm{\tau_1^s}$, $\bm{\tau_2^s}$ are color coded in red. For all time bumps, relation superscripts are dropped in the figure for better visibility.}
    \label{fig:overview}
\end{figure*}

\paragraph{The role of relation-specific representation.} BoxTE defines $k$ embeddings per time stamp $\tau$ (via the matrix $\mathbf{K^\tau}$) and $k$ scalars per relation $r$ (the vector $\bm{\alpha^r}$), which it linearly combines to compute time bumps. This allows individual relations to easily learn \emph{distinct} temporal behaviors, while maintaining \emph{information sharing}. To illustrate this, we consider two extreme scenarios. On one hand, if $k=1$, then relations can only vary the \emph{magnitude} of time bumps, which is restrictive. On the other hand, if relations are assigned their own learnable time bumps ($k=|R|, \bm{\alpha^r}$ is a one-hot encoding of $r$), then the overall model is overparametrized, and time bumps do not share parameters. Therefore, the current setup provides a trade-off using the hyper-parameter $k$, supporting both representational flexibility and efficient parameter sharing.

\paragraph{Illustrating the model.} We illustrate BoxTE with an example, shown in \Cref{fig:overview}. In this example, and throughout the paper, we  fix $\mathbf{T}=\{\tau_1, ..., \tau_{|T|}\}$ and refer to individual time stamps as $\tau_i$, to reflect natural temporal ordering.

\begin{example}
Consider the following temporal knowledge graph $\mathcal{G}=\{\temptriple{h}{r}{t}{\tau_1}, \temptriple{t}{s}{h}{\tau_1}, \temptriple{t}{s}{h}{\tau_2}\}$, defined over  ${\mathbf{E}=\{h, t\}}$, ${\mathbf{R}=\{r, s\}}$, and ${\mathbf{T}=\{\tau_1, \tau_2\}}$. The BoxTE configurations at time stamps $\tau_1$ and $\tau_2$ are shown on the left and right of \Cref{fig:overview}, respectively. The relations $r$ and $s$ are color coded with black and red, respectively. 

Observe that the fact $\triple{h}{r}{t}$ is true in time stamp $\tau_1$, but false at time stamp $\tau_2$, while the fact $\triple{t}{s}{h}$ remains true for both time stamps. 
In more detail, BoxTE assigns larger-magnitude time bumps $\bm{\tau_1^r}$ and  $\bm{\tau_2^r}$ for the relation $r$. In turn,  $\bm{\tau_1^r}$ makes $\temptriple{h}{r}{t}{\tau_1}$ true, as this bump moves the final representation of $h$ into $\bm{r^h}$, and  $\bm{\tau_2^r}$ makes $\temptriple{h}{r}{t}{\tau_2}$ false, by bumping this representation away from $\bm{r^h}$. By contrast, for the relation $s$, $\bm{\tau_1^s}$ and $\bm{\tau_2^s}$ are smaller, and do not ultimately affect the correctness of the fact $\triple{t}{s}{h}$ over time. As all time bumps are computed from the same matrices $\mathbf{K^{\tau_1}}$ and $\mathbf{K^{\tau_2}}$, this implies that BoxTE learns a smaller-norm $\bm{\alpha^s}$, reflecting the temporal stability of relation $s$.
\end{example}

\section{Model Properties}
We now study the representation power and inductive capacity of BoxTE. In particular, we show that BoxTE is fully expressive and extends the inductive capacity of BoxE to capture inference patterns \emph{across time}. 

\subsection{Full expressiveness} We first show that BoxTE is fully expressive, by proving the following statement:

\begin{theorem}
\label{thm:exp}
BoxTE is a fully expressive model for temporal knowledge graphs with the embedding dimensionality $d$ of entities, relations, bumps, and time bumps set to $d = \min ( \left| \mathbf{R} \right| \left| \mathbf{T} \right| \left| \mathbf{E} \right|,  \left| \mathbf{R} \right| \left| \mathbf{E} \right| ^2)$.
\end{theorem}

Importantly, this result already holds for the special case where $k=1$ and $\bm{\alpha^r}=1, \forall r\in\mathbf{R}$, i.e., for time bumps defined as relation-independent learnable embeddings. 

The full proof of this result can be found in the appendix. Briefly, the proof verifies two lemmas, corresponding to the $|\mathbf{R}||\mathbf{T}||\mathbf{E}|$ and $|\mathbf{R}||\mathbf{E}|^2$ bounds respectively, by providing two BoxTE constructions. For the former lemma, the construction builds on the original BoxE full expressiveness proof \cite{BoxE-NeurIPS20}, as bumps are used to make a given fact false without affecting other facts. However, this proof additionally relies on time bumps to make a temporal fact $\temptriple{h}{r}{t}{\tau_i}$ false without affecting the correctness of facts of the form $\temptriple{h}{r}{t}{\tau_j}$, where $j \neq i$. For the latter lemma, the constructions starts from a static BoxE configuration, and uses time bumps to make temporal facts \emph{true}, while using relation boxes to maintain the correctness and falsehood of other facts.

\subsection{Inference patterns} When studying inductive capacity, we say a model captures an inference pattern if it admits a set of parameters \emph{exactly} and \emph{exclusively} satisfying the pattern, following BoxE~\cite{BoxE-NeurIPS20}. 

\paragraph{Rigid inference patterns and relations.} BoxTE inherits the inductive capacity of BoxE in the static setting, where each pattern holds at \emph{all} time stamps. We refer to such patterns as \emph{rigid inference patterns}, as, for example, the rule:
\[
\forall \tau_i, \tau_j \in \mathbf{T}:
\big(\forall x,y~\temptriple{x}{r_1}{y}{\tau_i} \Rightarrow \temptriple{x}{r_2}{y}{\tau_j}\big)
\]
specifies that the relation $r_1$ is subsumed by the relation $r_2$ regardless of the time stamps we consider, and thus represents a rigid property. While we quantify over given time stamps $\mathbf{T}$, it is worthwhile noting that rigidity holds beyond known time stamps (i.e, the property extrapolates). 

Rigid inference patterns can be captured in BoxTE, as standard BoxE can be emulated by setting identical (or zero-valued) relation scalars, i.e., $\forall r, s \in \mathbf{R}, \bm{\alpha^r} = \bm{\alpha^s}$. Hence, BoxTE captures any inference pattern, and even rule language, captured by BoxE, in this sense.

Additionally, BoxTE can capture rigid relations, i.e., relations that do not vary with time, defined as:
\begin{align*}
\forall x,y~\big((\forall \tau \in \mathbf{T}: \temptriple{x}{r}{y}{\tau}) \lor (\forall \tau  \in \mathbf{T}: \neg \temptriple{x}{r}{y}{\tau})\big)
\end{align*}
This essentially implies that a relation $r$ is static over time, and corresponds to the parametrization $\bm{\alpha^r} = \bm{0}$.

\begin{table}[t]
\begin{tabular}{l}
\toprule
\textbf{Cross-time inference patterns}\\
\midrule
 Inversion:  $\temptriple{x}{r_1}{y}{\tau_1} \Leftrightarrow \temptriple{y}{r_2}{x}{\tau_2}$ \\
 Hierarchy:  $\temptriple{x}{r_1}{y}{\tau_1} \Rightarrow \temptriple{x}{r_2}{y}{\tau_2}$ \\
 Intersection:  $\temptriple{x}{r_1}{y}{\tau_1} \land \temptriple{x}{r_2}{y}{\tau_2} \Rightarrow \temptriple{x}{r_3}{y}{\tau_3}$\\
 Composition  $\temptriple{x}{r_1}{y}{\tau_1} \land \temptriple{y}{r_2}{z}{\tau_2} \Rightarrow \temptriple{x}{r_3}{z}{\tau_3}$\\
 Mutual exclusion: $\temptriple{x}{r_1}{y}{\tau_1} \land \temptriple{x}{r_2}{y}{\tau_2} \Rightarrow \bot$ \\
 \bottomrule
\end{tabular}
\caption{Cross-time inference patterns, where we omit universal quantification over variables. BoxTE captures all these patterns, but composition. Fixed-time inference patterns are the special case where the time stamps coincide.}
\label{tab:inf_pat}
\end{table}
\paragraph{Cross-time inference patterns.} In the temporal setting, a more interesting case is the study of inference patterns \emph{across specific time stamps}. For example, we may be interested in capturing the pattern for some fixed $\tau_1$, $\tau_2$:
\[
\forall x,y~\temptriple{x}{r_1}{y}{\tau_1} \Rightarrow \temptriple{x}{r_2}{y}{\tau_2}
\]
which requires $r_1$ at $\tau_1$ to be subsumed in $r_2$ at $\tau_2$, without any implications on the state of these relations on other time stamps.
The list of such cross-time inference patterns is provided in \Cref{tab:inf_pat}. BoxTE captures all these cross-time inference patterns, but composition.
To study cross-time patterns, we define, for a time stamp $\tau$ and relation $r$,  the time-induced relation head box $\bm{r^{h|\tau}}$ and the time-induced relation tail box $\bm{r^{t|\tau}}$ as:
\begin{align*}
\bm{r^{h|\tau}} = \bm{r^h} - \bm{\tau^r}, \quad\quad\quad \bm{r^{t|\tau}} = \bm{r^t} - \bm{\tau^r}.
\end{align*}

Intuitively, time-induced relation boxes offer an alternative, but equivalent, perspective to BoxTE representations, where time bumps apply to \emph{relation boxes} and translate them in the opposite direction. To show this equivalence, we recall the correctness criteria for BoxTE for a given fact $\temptriple{h}{r}{t}{\tau}$:
\begin{align*}
 \bm{e_h^{\temptriple{h}{r}{t}{\tau}}} \in \bm{r^h}, \quad\quad \bm{e_t^{\temptriple{h}{r}{t}{\tau}}} \in \bm{r^t}.
\end{align*}
Subtracting $\bm{\tau^r}$ from both sides yields:
\begin{align*}
&\bm{e_h^{\triple{h}{r}{t}}} \in \bm{r^{h|\tau}}, \quad\quad &&\bm{e_t^{\triple{h}{r}{t}}} \in \bm{r^{t|\tau}} .
\end{align*}
Hence, BoxTE can be interpreted as inducing, at every time stamp, a \emph{translated} set of relation boxes, which can then be compared with \emph{static} final entity embeddings to verify temporal fact correctness. Therefore, cross-time inference patterns can be studied by comparing time-induced relation boxes, analogously to standard BoxE. In particular, the \emph{hierarchy} pattern is captured with the parametrization:
\[
\bm{r_1^{h|\tau_1}} \subset \bm{r_2^{h|\tau_2}} \quad,  \quad\bm{r_1^{t|\tau_1}} \subset \bm{r_2^{t|\tau_2}},
\]
where $\subset$ denotes box containment. The \emph{intersection} pattern is captured by setting:
\[
\bm{r_3^{h|\tau_3}} = \bm{r_1^{h|\tau_1}} \cap \bm{r_2^{h|\tau_2}} \quad,  \quad \bm{r_3^{t|\tau_3}} = \bm{r_1^{t|\tau_1}} \cap \bm{r_2^{t|\tau_2}},
\]
where $\cap$ denotes box intersection. Mutual exclusion and inversion hold by analogous arguments. Note that capturing inference patterns at a \emph{fixed} time stamp is a special case.

By capturing cross-time inference patterns, BoxTE can model the interplay between relations across time. For instance, BoxTE can represent that any two entities engaging in formal negotiations at time stamp $\tau_1$ sign a formal agreement at time stamp $\tau_2$. It can also model relation behavior at single time stamp granularity, and thus capture patterns specifically at times when they hold, while learning appropriate temporal relation-specific behaviors. By contrast, existing models have shortcomings  when combining inference patterns and time modeling. For instance, TTransE can only capture rigid inverse relations if either the relation translation is set to $0$, or if the time translation is set to $0$, effectively eliminating the temporal component of the model.

\section{Experiments}
In this section, we evaluate BoxTE on TKG benchmarks ICEWS14, ICEWS15 \cite{garcia-duran-etal-2018-learning}, and GDELT \cite{Leetaru13gdelt:global}. We run experiments both in the standard temporal graph completion setting, and with the recently proposed bounded-parameter setting \cite{Lacroix2020Tensor}. We first present these datasets, and then report setup and results for both aforementioned experiments. 

In addition to these two experiments, we study the interpretability of BoxTE using a subset of YAGO \cite{YAGO_dataset}, perform an ablation study on BoxTE with different model variations, and conduct a robustness analysis relative to embedding dimensionality. These additional experiments can be found in the appendix.

\paragraph{Datasets.} 
We briefly present ICEWS14, ICEWS5-15, and GDELT, and report their main statistics in \Cref{tab:dataset_stats}. 
 \begin{table}[t!]
    \centering
    \begin{tabular}{llll}
        \toprule
                           & ICEWS14 & ICEWS05-15 & GDELT   \\
        \midrule
        $\left| \mathbf{E} \right|$ & 7,128   & 10,488     & 500        \\
        $\left| \mathbf{R} \right|$ & 230     & 251        & 20         \\
        $\left| \mathbf{T} \right|$ & 365     & 4017       & 366        \\
        $\left| \mathcal{G}_\text{train} \right|$            & 72,826  & 386,962    & 2,735,685  \\
        $\left| \mathcal{G}_\text{valid} \right|$            & 8,963   & 46,092     & 341,961    \\
        $\left| \mathcal{G}_\text{test} \right|$             & 8,941   & 46,275     & 341,961    \\
        $\left| \mathcal{G} \right|$            & 90,730  & 479,329    & 3,419,607   \\
        Timespan           & 1 year  & 11 years   & 1 year     \\
        Granularity        & Daily   & Daily      & Daily      \\
        \bottomrule
    \end{tabular}
    \caption{TKGC datasets with dataset statistics}
\label{tab:dataset_stats}
\end{table}
\begin{table*}[t] 
	\centering
	\label{tab:results_best} 
	\small\addtolength{\tabcolsep}{-1pt}
    \begin{tabular}{l@{\hskip 10pt}c@{\hskip 7pt}c@{\hskip 7pt}cc@{\hskip 7pt}c@{\hskip 10pt}c@{\hskip 7pt}c@{\hskip 7pt}cc@{\hskip 7pt}c@{\hskip 10pt}c@{\hskip 7pt}c@{\hskip 7pt}cc@{\hskip 7pt}c@{\hskip 7pt}}  %
		\toprule 
		 {Model} & \multicolumn{5}{c}{\textbf{ICEWS14}} & \multicolumn{5}{c}{\textbf{ICEWS5-15}} & \multicolumn{5}{c}{\textbf{GDELT}} \\
		\cmidrule(r){2-6}
		\cmidrule(r){7-11}
		\cmidrule(r){12-16}
		 & MR & MRR & H@1 & H@3 & H@10 & MR & MRR & H@1 & H@3 & H@10 & MR & MRR & H@1 & H@3 & H@10\\
		 TTransE & - & .255 & .074 & - & .601 & - & .271 & .084 & - & .616 & - & .115 & 0.0 & .160 & .318 \\
		 DE-SimplE & - & .526 & .418 & .592 & .725 & - & .513 & .392 & .578 & .748 & - & .230 & .141 & .248 & .403\\
		 TA-DistMult & - & .477 & .363 & - & .686 & - & .474 & .346 & - & .728 & - & .206 & .124 & .219 & .365\\
		 ChronoR(a) & - & .594 & .496 & .654 & .773 & - & .684 & \textbf{.611} & .730 & .821 & - & - & - & - & -\\
		 ChronoR(b) & - & \textbf{.625} & \textbf{.547} & .669 & .773 & - & .675 & .596 & .723 & .820 & - & - & - & - & -\\
		 \midrule 
		 TComplEx & - & .610 & - & - & - & - & .660 & - & - & - & - & - & - & - & -\\
		 TNTComplEx & - & .620 & .520 & .660 & .760 & - & .670 & .590 & .710 & .810 & - & - & - & - & -\\
		 TeLM & - & \textbf{.625} & .545 & .673 & .774 & - & .678 & .599 & .728 & .823 & - &  - & - & -  & -\\
		 \midrule 
		 TeMP-SA & - &.607 & .484 & \textbf{.684} & \textbf{.840} & - & .680 & .553 & .769 &.913 & - & .232 & .152 & .245 & .377 \\
		 TeMP-GRU & - & .601 & .478 & .681 & .825 & - & \textbf{.691} & .566 & \textbf{.782} & \textbf{.917} & - & .275 & .191 & .297 & .437\\
		 \midrule
		 BoxTE~(k=2) & 161 & .615 & .532 & .667 & .767 & 98 & .664 & .576 & .720 & .822 & 48 & .339 & .251 &  .366 & .507\\
		 BoxTE~(k=3) & 162 & .614 & .530 & .668 & .765 & 101 & .666 & .582 & .719 & .820 & 49 & .344 & .259 &  .369 & .507\\
		 BoxTE~(k=5) & 160 & .613 & .528 & .664 & .763 & 96 & .667 & .582 & .719 & .820 & 50 & \textbf{.352} & \textbf{.269} &  \textbf{.377} & \textbf{.511} \\
		\bottomrule
	\end{tabular}
	\caption{TKGC results for BoxTE on ICEWS14, ICEWS5-15, and GDELT. Results for competing models are the best reported from their respective papers, which are referenced in \Cref{sec:rw}. ChronoR(a) uses three-dimensional rotation, whereas ChronoR(b) uses two-dimensional rotation. } 
	\label{tab:results_tkgc}
\end{table*}
ICEWS14 and ICEWS5-15 \cite{garcia-duran-etal-2018-learning} are both subsets of the \emph{Integrated Crisis Early Warning System (ICEWS)} dataset \cite{ICEWS_dataset}, which stores temporal socio-political facts starting from the year 1995. More specifically, ICEWS14 includes facts from the year 2014 involving frequently occurring entities, and ICEWS5-15 includes analogous facts from the period between 2005 and 2015 inclusive. %

\noindent GDELT is a subset of the larger \emph{Global Database of Events, Language, and Tone} (GDELT) temporal knowledge graph \cite{Leetaru13gdelt:global}, which stores facts about human behavior starting from 1979. This benchmark contains facts with time stamps between April 1, 2015 and March 31, 2016, and only includes facts involving the 500 most frequent entities and 20 most common relations. 
 
\subsection{Temporal knowledge graph completion}
\paragraph{Experimental setup.} In this experiment, we train BoxTE on all three benchmark datasets, report test set performance for the best validation setup and compare against baseline models for TKGC. More specifically, we evaluate BoxTE in terms of mean rank (MR), mean reciprocal rank (MRR), Hits@1, Hits@3, and Hits@10, and tune its embedding dimensionality $d$, training batch sizes, number of negative samples, and $k$ values within the set $\{2, 3, 5\}$. We train BoxTE with cross-entropy loss, and use the Adam optimizer \cite{kingma2017adam} with the default learning rate of $10^{-3}$. We additionally conduct experiments using the temporal smoothness regularizer proposed in TNTComplEx \cite{Lacroix2020Tensor}, and consider factorizations of time embeddings to encourage parameter sharing. Full details about the hyper-parameter setup used to obtain results and explanations of these hyper-parameters are provided in the appendix. 

\paragraph{Results.} Results for the standard TKGC setting are shown in \Cref{tab:results_tkgc}. BoxTE achieves state-of-the art performance on GDELT, comfortably surpassing TeMP \cite{wu2020temp} in terms of MRR. BoxTE also performs strongly on ICEWS14, and ICEWS15. On ICEWS14, BoxTE also outperforms TeMP, and is competitive with TNTComplEx and ChronoR. %
This trend also carries to ICEWS15, where BoxTE remains strong despite the sparsity of the dataset.

These results are highly encouraging, particularly given the difficulty and size of the GDELT dataset. Indeed, GDELT is substantially denser than both ICEWS datasets (${\sim2.7}$ million training facts for 500 entities, 20 relations), and involves significant temporal variability. More concretely, some facts persist across multiple consecutive time stamps, whereas others are momentary and sparse. Hence, GDELT requires strong inductive capacity, and thus is highly challenging. In fact, most TKGC models fail to beat the simple rule-based system TED \cite{wu2020temp} on GDELT.  Hence, the very strong performance of BoxTE suggests that the model does capture temporal patterns, and hence exploits information in the data to outperform existing models. 

\begin{table*}[t] 
	\centering
	\small\addtolength{\tabcolsep}{-1pt}
	\resizebox{\textwidth}{!}{\begin{tabular}{l@{\hskip 10pt}c@{\hskip 7pt}c@{\hskip 7pt}cc@{\hskip 7pt}c@{\hskip 10pt}c@{\hskip 7pt}c@{\hskip 7pt}cc@{\hskip 7pt}c@{\hskip 10pt}c@{\hskip 7pt}c@{\hskip 7pt}cc@{\hskip 7pt}c@{\hskip 7pt}}
		\toprule 
		 {Model} & \multicolumn{5}{c}{\textbf{ICEWS14}} & \multicolumn{5}{c}{\textbf{ICEWS5-15}} & \multicolumn{5}{c}{\textbf{GDELT}} \\
		\cmidrule(r){2-6}
		\cmidrule(r){7-11}
		\cmidrule(r){12-16}
		 & MR & MRR & H@1 & H@3 & H@10 & MR & MRR & H@1 & H@3 & H@10 & MR & MRR & H@1 & H@3 & H@10\\
		 DE-SimplE & - & .526 & .418 & .592 & .725 & - & .513 & .392 & .578 & .748 & - & .230 & .141 & .248 & .403\\
		 TComplEx & - & .560 & - & - & - & - & .580 & - & - & - & - & - & - & - & -\\
		 TNTComplEx & - & .560 & - & - & - & - & \textbf{.600} & - & - & - & - & - & - & - & -\\
		 \midrule
		 BoxTE~(k=1) & 183 & .576 & .478 & .639 & .753 & 122 & .564 & .452 & .635 & .770 & 62 & \textbf{.250} & \textbf{.167} &  \textbf{.270} & \textbf{.411}\\
		 BoxTE~(k=2) & 177 & .580 & .483 & \textbf{.642} & \textbf{.755} & 110 & .567 & .458 & .631 & \textbf{.775} & 63 & .246 & .164 &  .265 & .404\\
		 BoxTE~(k=3) & 182 & \textbf{.582} & .491 & .640 & .748 & 125 & .570 & .465 & \textbf{.636} & .763 & 64 & .242 & .161 &  .260 & .398\\
		 BoxTE~(k=5) & 183 & .581 & \textbf{.493} & .632& .742 & 134 & .567 & \textbf{.469} & .623 & .746 & 66 & .236 & .156 &  .253 & .390 \\
		\bottomrule
	\end{tabular}}
    	\caption{TKGC results for BoxTE and competing models in the bounded-parameter setting.}%
	\label{tab:results_param_match} 
\end{table*}

On the other hand, both ICEWS datasets are small and, critically, very sparse. For example, ICEWS14 only includes around 70k training facts for 7,000 entities across 365 timestamps. Furthermore, the relations in these datasets are sparse, as they usually encode one-time patterns with limited, if any, regularity, e.g., official visits, negotiations, statements. Hence, these datasets include substantially less temporal patterns and variability, and rely more on entity-driven predictions. This is further highlighted by the fact that the same TKGC models performing poorly on GDELT substantially outperform the rule-based system TED on ICEWS14 and ICEWS5-15 \cite{wu2020temp}. Hence, both datasets include few temporal patterns for BoxTE to capture, and this substantially reduces the inductive advantage of this model relative to its competitors.  

In terms of performance relative to $k$, BoxTE performs best with  $k=2$ on ICEWS14, whereas the optimal $k$ value is 5 for both ICEWS5-15 and GDELT. This may seem unintuitive, especially given the large similarity between both ICEWS datasets, but this can be attributed to the significantly larger number of time steps in ICEWS5-15. Indeed, ICEWS5-15 includes 4017 time stamps, whereas ICEWS14 only includes 365. Thus, more flexibility is needed to learn sufficiently distinct temporal behaviors across these time stamps, and this aligns with our intuition about the advantages of higher $k$. GDELT results also highlight the importance of higher $k$, as they confirm the need for more flexibility to fit the rich set of facts it provides.

Finally, we note that BoxTE is very robust when training on these benchmarks, as it maintains strong performance even when not supplemented with temporal smoothness regularization. By contrast, models such as TNTComplEx and ChronoR suffer significantly without regularization. This further highlights the inductive capacity of BoxTE, which can autonomously learn temporal properties from data, and suggests that this model is a strong, natural choice for applications on novel datasets where such regularizations are not known or not compatible with the data. We discuss this experiment in more detail as an ablation study in the appendix. 

\subsection{Parameter-bounded experiments for TKGC}

\paragraph{Experimental setup.} In this experiment, we train and evaluate BoxTE analogously to the standard TKGC setup, but, we impose a dimensionality constraint, such that the total number of parameters used by the model does not exceed that of 100-dimensional DE-SimplE, in keeping with the literature \cite{Lacroix2020Tensor}. Hence, this experiment aims to evaluate the robustness of models with a restricted computational budget. Given this restriction, we additionally evaluate BoxTE with $k=1$, so as to allow for a slightly improved dimensionality, albeit at the expense of  flexibility. Further details about the parameter counts of BoxTE and competing models are provided in the appendix.

\paragraph{Results.} Results for the bounded-parameter setup are reported in Table \ref{tab:results_param_match}. In this setup, we see that BoxTE now achieves state-of-the-art performance on both ICEWS14 and GDELT, and maintains its strong performance despite parametrization constraints. In particular, BoxTE only drops by 0.03 in terms of MRR on ICEWS14 relative to its performance on in the standard setup, whereas it drops by 0.10 on GDELT, and by 0.09 on ICEWS5-15. This further reflects the simplicity of ICEWS14, as a low number of parameters remains sufficient for high performance, and highlights the richness of GDELT, as well as the complexity of ICEWS5-15, owing primarily to its larger number of time stamps. 

The results of this experiment relative to different $k$ also portray an interesting interplay between this parameter and embedding dimensionality $d$, which manifests differently across the three benchmark datasets. Conceptually, the bounded-parameter setting imposes a trade-off on the choice of $k$: a small $k$ maximizes dimensionality, but reduces flexibility, whereas a larger $k$ is more flexible, but substantially reduces the available dimensionality. Hence, the optimal value of $k$ is not obvious, and varies substantially among datasets. On ICEWS14, we see that $k$ values of 2 and above perform similarly well, and slightly outperform the $k=1$ model, despite their lower dimensionality. This also aligns with the optimal $k$ in the standard setting, and suggests that capturing relational temporal dynamics to some extent (via $k > 1$) on ICEWS14 is more important than storing more information through larger dimensionality. 

On ICEWS5-15, we see that MRR improves as $k$ increases, and that $k=3$ provides strong overall performance. This improvement relative to $k$ is surprising, as the large number of time stamps in this dataset causes embedding dimensionality to decrease substantially as $k$ increases. In fact, $d=137$ when $k=1$, but this drops to $d=104$ when $k=3$. Nonetheless, model performance improves. This further highlights the higher importance of capturing relational temporal dynamics in ICEWS5-15 relative to higher dimensionality, and aligns with our expectations given the large number of time stamps in this dataset. 

Finally, for GDELT, BoxTE achieves optimal performance with only $k=1$, which stands in sharp contrast with $k=5$ being optimal in the standard setting. However, this can be traced back to the density of this dataset. Indeed, as every entity and relation appears in a large number of facts, more dimensionality is needed to capture the information represented by existing facts. Hence, the reduction of dimensionality in this setup causes significant loss in representation capacity, and this severely hurts BoxTE on GDELT. Given this bottleneck, increasing dimensionality is substantially more beneficial in the parameter-bounded setting than improving flexibility, and thus lowering $k$ offers more gain on GDELT. 

Overall, these results show that BoxTE offers a state-of-the-art, robust baseline for temporal knowledge base completion even under a restricted computational budget.   

\section{Summary and Outlook}
In this paper, we presented BoxTE, a temporal knowledge graph embedding model, and showed that this model is fully expressive and captures a rich class of temporal inference patterns. We then evaluated BoxTE empirically, and showed that the model achieves state-of-the-art performance for TKGC, and benefits substantially from its inductive capacity and robustness. 
Similarly to BoxE, BoxTE naturally applies to higher-arity knowledge bases. Unfortunately, there are no established benchmarks for higher-arity temporal knowledge graph completion, despite its significant potential and wide applicability. One interesting future direction is therefore, introducing new benchmarks for temporal knowledge base completion, involving higher-arity facts to study the performance of BoxTE, as well as other models, in this setting. 
We think that this work will motivate further research leading to the development of expressive, inductively rich TKGE models.

\bibliography{aaai22}

\begin{thebibliography}{30}
\providecommand{\natexlab}[1]{#1}

\bibitem[{Abboud et~al.(2020)Abboud, Ceylan, Lukasiewicz, and
  Salvatori}]{BoxE-NeurIPS20}
Abboud, R.; Ceylan, {\.I}.~{\.I}.; Lukasiewicz, T.; and Salvatori, T. 2020.
\newblock Box{E}: {A} Box Embedding Model for Knowledge Base Completion.
\newblock In \emph{Proceedings of the Thirty-Fourth Annual Conference on
  Advances in Neural Information Processing Systems, {NeurIPS}}.

\bibitem[{Balazevic, Allen, and Hospedales(2019)}]{TuckER}
Balazevic, I.; Allen, C.; and Hospedales, T.~M. 2019.
\newblock Tuck{ER}: Tensor Factorization for Knowledge Graph Completion.
\newblock In \emph{Proceedings of the 2019 {C}onference on {E}mpirical
  {M}ethods in {N}atural {L}anguage {P}rocessing and the {N}inth
  {I}nternational {J}oint {C}onference on {N}atural {L}anguage {P}rocessing,
  {EMNLP-IJCNLP}}, 5184--5193.

\bibitem[{Bollacker, Cook, and Tufts(2007)}]{BollackerCT07}
Bollacker, K.~D.; Cook, R.~P.; and Tufts, P. 2007.
\newblock Freebase: {A} Shared Database of Structured General Human Knowledge.
\newblock In \emph{Proceedings of the Twenty-Second {AAAI} Conference on
  Artificial Intelligence, {AAAI}}.

\bibitem[{Bordes, Chopra, and Weston(2014)}]{BordesCW14}
Bordes, A.; Chopra, S.; and Weston, J. 2014.
\newblock Question Answering with Subgraph Embeddings.
\newblock In \emph{Proceedings of the 2014 {C}onference on {E}mpirical
  {M}ethods in {N}atural {L}anguage {P}rocessing, {EMNLP}}.

\bibitem[{Bordes et~al.(2013)Bordes, Usunier, Garc{\'{\i}}a{-}Dur{\'{a}}n,
  Weston, and Yakhnenko}]{TransE-NIPS13}
Bordes, A.; Usunier, N.; Garc{\'{\i}}a{-}Dur{\'{a}}n, A.; Weston, J.; and
  Yakhnenko, O. 2013.
\newblock Translating Embeddings for Modeling Multi-relational Data.
\newblock In \emph{Proceedings of the {T}wenty-{S}ixth {A}nnual {C}onference on
  {A}dvances in {N}eural {I}nformation {P}rocessing {S}ystems, {NIPS}},
  2787--2795.

\bibitem[{Boschee et~al.(2015)Boschee, Lautenschlager, O'Brien, Shellman,
  Starz, and Ward}]{ICEWS_dataset}
Boschee, E.; Lautenschlager, J.; O'Brien, S.; Shellman, S.; Starz, J.; and
  Ward, M. 2015.
\newblock {ICEWS Coded Event Data}.

\bibitem[{Dasgupta, Ray, and Talukdar(2018)}]{HyTE}
Dasgupta, S.~S.; Ray, S.~N.; and Talukdar, P. 2018.
\newblock {H}y{TE}: Hyperplane-based Temporally aware Knowledge Graph
  Embedding.
\newblock In \emph{Proceedings of the 2018 Conference on Empirical Methods in
  Natural Language Processing , {EMNLP}}, 2001--2011. Association for
  Computational Linguistics.

\bibitem[{Dettmers et~al.(2018)Dettmers, Minervini, Stenetorp, and
  Riedel}]{ConvE-AAAI18}
Dettmers, T.; Minervini, P.; Stenetorp, P.; and Riedel, S. 2018.
\newblock Convolutional 2{D} Knowledge Graph Embeddings.
\newblock In \emph{Proceedings of the Thirty-Second {AAAI} Conference on
  Artificial Intelligence, {AAAI}}, 1811--1818.

\bibitem[{Dong et~al.(2014)Dong, Gabrilovich, Heitz, Horn, Lao, Murphy,
  Strohmann, Sun, and Zhang}]{GoogleVault}
Dong, X.; Gabrilovich, E.; Heitz, G.; Horn, W.; Lao, N.; Murphy, K.; Strohmann,
  T.; Sun, S.; and Zhang, W. 2014.
\newblock {Knowledge Vault: A} web-scale approach to probabilistic knowledge
  fusion.
\newblock In \emph{Proceedings of the Twentieth {ACM} {SIGKDD} International
  Conference on Knowledge Discovery and Data Mining, KDD}.

\bibitem[{Garc{\'\i}a-Dur{\'a}n, Duman{\v{c}}i{\'c}, and
  Niepert(2018)}]{garcia-duran-etal-2018-learning}
Garc{\'\i}a-Dur{\'a}n, A.; Duman{\v{c}}i{\'c}, S.; and Niepert, M. 2018.
\newblock Learning Sequence Encoders for Temporal Knowledge Graph Completion.
\newblock In \emph{Proceedings of the 2018 Conference on Empirical Methods in
  Natural Language Processing , {EMNLP}}, 4816--4821. Association for
  Computational Linguistics.

\bibitem[{Goel et~al.(2019)Goel, Kazemi, Brubaker, and
  Poupart}]{DiachronicEmbedding}
Goel, R.; Kazemi, S.~M.; Brubaker, M.; and Poupart, P. 2019.
\newblock Diachronic Embedding for Temporal Knowledge Graph Completion.
\newblock \emph{CoRR}, abs/1907.03143.

\bibitem[{Kazemi et~al.(2020)Kazemi, Goel, Jain, Kobyzev, Sethi, Forsyth, and
  Poupart}]{KazemiGJKSFP20}
Kazemi, S.~M.; Goel, R.; Jain, K.; Kobyzev, I.; Sethi, A.; Forsyth, P.; and
  Poupart, P. 2020.
\newblock Representation Learning for Dynamic Graphs: {A} Survey.
\newblock \emph{J. Mach. Learn. Res.}, 21: 70:1--70:73.

\bibitem[{Kingma and Ba(2017)}]{kingma2017adam}
Kingma, D.~P.; and Ba, J. 2017.
\newblock Adam: A Method for Stochastic Optimization.
\newblock arXiv:1412.6980.

\bibitem[{Lacroix, Obozinski, and Usunier(2020)}]{Lacroix2020Tensor}
Lacroix, T.; Obozinski, G.; and Usunier, N. 2020.
\newblock Tensor Decompositions for Temporal Knowledge Base Completion.
\newblock In \emph{Proceedings of the Eighth International Conference on
  Learning Representations ({ICLR})}.

\bibitem[{Leblay and Chekol(2018)}]{TTransE}
Leblay, J.; and Chekol, M.~W. 2018.
\newblock Deriving Validity Time in Knowledge Graph.
\newblock In Champin, P.-A.; Gandon, F.; Lalmas, M.; and Ipeirotis, P.~G.,
  eds., \emph{Proceedings of the International World Wide Web Conference ({
  WWW})}, 1771--1776. {ACM Press}.
\newblock ISBN 9781450356404.

\bibitem[{Leetaru and Schrodt(2013)}]{Leetaru13gdelt:global}
Leetaru, K.; and Schrodt, P.~A. 2013.
\newblock GDELT: Global data on events, location, and tone.
\newblock \emph{ISA Annual Convention}.

\bibitem[{Mahdisoltani, Biega, and Suchanek(2015)}]{MahdisoltaniBS15}
Mahdisoltani, F.; Biega, J.; and Suchanek, F.~M. 2015.
\newblock {YAGO3:} {A} Knowledge Base from Multilingual Wikipedias.
\newblock In \emph{Proceedings of the Seventh Biennial Conference on Innovative
  Data Systems Research, {CIDR}}.

\bibitem[{Mitchell et~al.(2018)Mitchell, Cohen, Jr., Talukdar, Yang,
  Betteridge, Carlson, Mishra, Gardner, Kisiel, Krishnamurthy, Lao, Mazaitis,
  Mohamed, Nakashole, Platanios, Ritter, Samadi, Settles, Wang, Wijaya, Gupta,
  Chen, Saparov, Greaves, and Welling}]{MitchellBCM18}
Mitchell, T.~M.; Cohen, W.~W.; Jr., E. R.~H.; Talukdar, P.~P.; Yang, B.;
  Betteridge, J.; Carlson, A.; Mishra, B.~D.; Gardner, M.; Kisiel, B.;
  Krishnamurthy, J.; Lao, N.; Mazaitis, K.; Mohamed, T.; Nakashole, N.;
  Platanios, E.~A.; Ritter, A.; Samadi, M.; Settles, B.; Wang, R.~C.; Wijaya,
  D.; Gupta, A.; Chen, X.; Saparov, A.; Greaves, M.; and Welling, J. 2018.
\newblock Never-ending learning.
\newblock \emph{Communications of the {ACM}}, 61(5): 103--115.

\bibitem[{Nickel, Tresp, and Kriegel(2011)}]{RESCAL-ICML11}
Nickel, M.; Tresp, V.; and Kriegel, H. 2011.
\newblock A Three-Way Model for Collective Learning on Multi-Relational Data.
\newblock In \emph{Proceedings of the {T}wenty-{E}ighth {I}nternational
  {C}onference on {M}achine {L}earning, {ICML}}, 809--816.

\bibitem[{Sadeghian et~al.(2021)Sadeghian, Armandpour, Colas, and
  Wang}]{ChronoR}
Sadeghian, A.; Armandpour, M.; Colas, A.; and Wang, D.~Z. 2021.
\newblock ChronoR: Rotation Based Temporal Knowledge Graph Embedding.
\newblock In \emph{Proceedings of the Thirty-Fifth {AAAI} Conference on
  Artificial Intelligence ({AAAI})}, 6471--6479. {AAAI} Press.

\bibitem[{Schlichtkrull et~al.(2018)Schlichtkrull, Kipf, Bloem, van~den Berg,
  Titov, and Welling}]{SchlichtkrullKB18}
Schlichtkrull, M.~S.; Kipf, T.~N.; Bloem, P.; van~den Berg, R.; Titov, I.; and
  Welling, M. 2018.
\newblock Modeling Relational Data with Graph Convolutional Networks.
\newblock In \emph{Proceedings of the {F}ifteenth {I}nternational {C}onference
  on the {S}emantic {W}eb, {ESWC}}, 593--607.

\bibitem[{Suchanek, Kasneci, and Weikum(2007)}]{YAGO_dataset}
Suchanek, F.~M.; Kasneci, G.; and Weikum, G. 2007.
\newblock Yago: A Core of Semantic Knowledge.
\newblock In \emph{Proceedings of the Sixteenth International Conference on
  World Wide Web}, WWW, 697–706. Association for Computing Machinery.
\newblock ISBN 9781595936547.

\bibitem[{Sun et~al.(2019)Sun, Deng, Nie, and Tang}]{RotatE-ICLR19}
Sun, Z.; Deng, Z.; Nie, J.; and Tang, J. 2019.
\newblock Rotat{E}: Knowledge Graph Embedding by Relational Rotation in Complex
  Space.
\newblock In \emph{Proceedings of the {S}eventh {I}nternational {C}onference on
  {L}earning {R}epresentations, {ICLR}}.

\bibitem[{Toutanova and Chen(2015)}]{FB15k237TC}
Toutanova, K.; and Chen, D. 2015.
\newblock Observed versus latent features for knowledge base and text
  inference.
\newblock In \emph{Proceedings of the Third Workshop on Continuous Vector Space
  Models and their Compositionality}, 57--66. Association for Computational
  Linguistics.

\bibitem[{Trouillon et~al.(2016)Trouillon, Welbl, Riedel, Gaussier, and
  Bouchard}]{ComplEx}
Trouillon, T.; Welbl, J.; Riedel, S.; Gaussier, {\'{E}}.; and Bouchard, G.
  2016.
\newblock Complex Embeddings for Simple Link Prediction.
\newblock In \emph{Proceedings of the {T}hirty-{T}hird {I}nternational
  {C}onference on {M}achine {L}earning, {ICML}}, 2071--2080.

\bibitem[{Wang et~al.(2018)Wang, Zhang, Wang, Zhao, Li, Xie, and
  Guo}]{WangZWZLXG18}
Wang, H.; Zhang, F.; Wang, J.; Zhao, M.; Li, W.; Xie, X.; and Guo, M. 2018.
\newblock RippleNet: Propagating User Preferences on the Knowledge Graph for
  Recommender Systems.
\newblock In \emph{Proceedings of the Twenty-Seventh {ACM} International
  Conference on Information and Knowledge Management, {CIKM}}.

\bibitem[{Wu et~al.(2020)Wu, Cao, Cheung, and Hamilton}]{wu2020temp}
Wu, J.; Cao, M.; Cheung, J. C.~K.; and Hamilton, W.~L. 2020.
\newblock Te{MP}: Temporal Message Passing for Temporal Knowledge Graph
  Completion.
\newblock arXiv:2010.03526.

\bibitem[{Xiong, Power, and Callan(2017)}]{XiongPC17}
Xiong, C.; Power, R.; and Callan, J. 2017.
\newblock Explicit Semantic Ranking for Academic Search via Knowledge Graph
  Embedding.
\newblock In \emph{Proceedings of the Twenty-Sixth International Conference on
  World Wide Web, {WWW}}.

\bibitem[{Xu et~al.(2021)Xu, Chen, Nayyeri, and
  Lehmann}]{xu-etal-2021-temporal}
Xu, C.; Chen, Y.-Y.; Nayyeri, M.; and Lehmann, J. 2021.
\newblock Temporal Knowledge Graph Completion using a Linear Temporal
  Regularizer and Multivector Embeddings.
\newblock In \emph{Proceedings of the 2021 Conference of the North American
  Chapter of the Association for Computational Linguistics: Human Language
  Technologies}, 2569--2578. Online: Association for Computational Linguistics.

\bibitem[{Xu et~al.(2020)Xu, Nayyeri, Alkhoury, Yazdi, and
  Lehmann}]{xu2020tero}
Xu, C.; Nayyeri, M.; Alkhoury, F.; Yazdi, H.~S.; and Lehmann, J. 2020.
\newblock TeRo: {A} Time-aware Knowledge Graph Embedding via Temporal Rotation.
\newblock In Scott, D.; Bel, N.; and Zong, C., eds., \emph{Proceedings of the
  Twenty-Eighth International Conference on Computational Linguistics
  ({COLING})}, 1583--1593. International Committee on Computational
  Linguistics.

\end{thebibliography}

\appendix
\section{Proof of \Cref{thm:exp}}
We first recall the theorem as stated above:

\smallskip
\noindent\textbf{\Cref{thm:exp}}. \emph{BoxTE is a fully expressive model for temporal knowledge graphs with the embedding dimensionality $d$ of entities, relations, bumps, and time bumps set to $d = \min ( \left| \mathbf{R} \right| \left| \mathbf{T} \right| \left| \mathbf{E} \right|,  \left| \mathbf{R} \right| \left| \mathbf{E} \right| ^2)$.}
\smallskip

We prove this result by independently showing that BoxTE is fully expressive with $d=\left| \mathbf{R} \right| \left| \mathbf{T} \right| \left| \mathbf{E} \right|$ and $d=\left| \mathbf{R} \right| \left| \mathbf{E} \right| ^2$ in the following lemmas, which together imply \Cref{thm:exp}.

\begin{lemma} \label{lem:RTE}
BoxTE is a fully expressive model for temporal knowledge graphs with the embedding dimensionality $d$ of entities, relations, bumps, and time bumps set to $d = \left| \mathbf{R} \right| \left| \mathbf{T} \right| \left| \mathbf{E} \right|$.
\end{lemma}

\begin{lemma} \label{lem:REE}
BoxTE is a fully expressive model for temporal knowledge graphs with the embedding dimensionality $d$ of entities, relations, bumps, and time bumps set to $d = \left| \mathbf{R} \right| \left| \mathbf{E} \right| ^ 2$.
\end{lemma}

In what follows, we now identify relations, entities and timestamps using indices. More concretely, we consider relations $r_j \in \mathbf{R}$, entities $e_i \in \mathbf{E}$, time stamps $\tau_l \in \mathbf{T}$, as these indices make the proof more convenient.

\subsection{Proof of \Cref{lem:RTE}}
The proof of this lemma is an adaptation of the BoxE full expressiveness proof \cite{BoxE-NeurIPS20} and shows how, starting from a configuration that makes all possible facts true, arbitrary facts can be made false by bumping entities out of their target relation boxes.
We show by induction that any temporal fact in a graph $\mathcal{G}$ can be made false, while preserving the truth value of all other facts.

For a vector $\bm{v} \in \mathbb{R}^d$ we write $\bm{v}(j, k, l)$ to refer to the index $j \left| \mathbf{E} \right| \left| \mathbf{T} \right| + k \left| \mathbf{T} \right| + l$.
Intuitively, we assign a unique dimension in our embedding space to any entity-relation-time triple.
We denote the upper and lower boundaries of a box $\bm{r_j^x}$, $x \in \{h, t\}$ with $\bm{u_j^{x}}$ and $\bm{l_j^{x}}$, respectively.
Further, without loss of generality, we assume that each time stamp is represented by a single temporal bump ($k=1$), and that all temporal scalars are fixed to a value of one ($\alpha^{r_1}= ... = \alpha^{r_{\left| \mathcal{R} \right|}} = 1$).
In this setting, the single learned time bump for any time stamp directly represents the final time bump for any relation at that snapshot, simplifying the proof.
For time $\tau_i$, we denote this time bump with $\bm{\tau_{i}}$. 
The more general setup with multriple time bumps per snapshot and learnable temporal scalars can be configured to obtain the setting described above, thus making the following proof valid for general BoxTE. 

\paragraph{Base case:}
We initialize a BoxTE configuration that captures the complete universe of facts over the vocabulary of $\mathcal{G}$, meaning that all possible facts are set to be $true$.
This can be achieved by setting all entity bases, entity bumps, and time bumps to $\bm{0}$, and all relation boxes as unit boxes centred at $\bm{0}$.

\paragraph{Induction step:}
In this step we consider any arbitrary fact $\temptriple{e_i}{r_j}{e_k}{\tau_l}$ and make it false without affecting any other fact in the graph.
This can be achieved via the following steps:
\begin{itemize}
    \item[] \textbf{Step 1.} Increment $\bm{b_i}(j, k, l)$ by a value C, such that:
        
        \[
        \bm{e_k}(j, k, l) + \bm{b_i}(j, k, l) + \bm{\tau_{l}}(j, k, l) + C \quad > \quad \bm{u_j^{t}}(j, k, l)
        \]
    \item[] \textbf{Step 2.} Decrement all entity embeddings except $\bm{e_k}$ by C at dimension $(j, k, l)$:
        \[
        \forall k' \neq k : \bm{e_{k'}} (j, k, l) := \bm{e_{k'}} (j, k, l) - C
        \]
    \item[] \textbf{Step 3.} Decrement all time bumps except $\bm{\tau_{l}}$ by $C$ in dimension $(j, k, l)$:
        \[
        \forall l' \neq l : \bm{e_{k'}} (j, k, l) := \bm{\tau_{{l'}}} (j, k, l) - C
        \]
    \item[] \textbf{Step 4.} For relation $r_j$, grow the head box at dimension $(j, k, l)$ upwards by $C$ and downwards by $2C$, and grow the tail box downwards by $2C$ at the same dimension:
        \[
        \bm{l^{h}_j} (j, k, l) := \bm{l^{h}_j} (j, k, l) - 2C
        \]
        \[
        \bm{u^{h}_j} (j, k, l) := \bm{u^{h}_j} (j, k, l) + C
        \]
        \[
        \bm{l^{t}_j} (j, k, l) := \bm{l^{t}_j} (j, k, l) - 2C
        \]
    \item[] \textbf{Step 5.} For all other relations $r_x \in \mathcal{R}, x \neq j$, grow both boxes at dimension $(j, k, l)$, upwards by $C$ and downwards by $2C$:
    \[
        \bm{u^{h/t}_x} (j, k, l) := \bm{u^{h/t}_x} (j, k, l) + C
        \]
        \[
        \bm{l^{h/t}_x} (j, k, l) := \bm{l^{h/t}_x} (j, k, l) - 2C
        \]
\end{itemize}

It remains to be shown that the induction step does indeed flip the truth value of fact $\temptriple{e_i}{r_j}{e_k}{\tau_l}$ to false, while maintaining the truth value of all other facts.

Observe that Step 1 pushes $\bm{e_k^{\temptriple{e_i}{r_j}{e_k}{\tau_l}}}$ outside of its target relation box $\bm{r_j^{t}}$ at dimension $(j, k, l)$, thus making the fact false, as intended.

To show that the truth value of any other fact is unaffected by the induction step, we analyse the effect of steps 1-5 on any possible fact $F' = \temptriple{e_{i'}}{r_{j'}}{e_{k'}}{\tau_{l'}} \neq \temptriple{e_i}{r_j}{e_k}{\tau_l}$.
We do this by considering the head an tail entities in the following cases:

\begin{itemize}
    \item [] Case 1. \textbf{The fact $F'$ is true}: We show that this fact remains true after the inductive step by analysing both the head entity $\bm{e_{i'}^{F'}}$ and tail entity $\bm{e_{k'}^{F'}}$.
    \begin{enumerate}[label=(\alph*)]
        \item \textbf{Head entity:} The final embedding $\bm{e_{i'}^{F'}}$ can move by at most $C$ in the upwards direction and $2C$ in the downwards direction after steps 1 and 2, and all relation boxes are grown by $C$ and $2C$, respectively, after steps 3 and 4, so $\bm{e_{i'}^{F'}} \in \bm{r_{j'}^{h}}$ still holds.
        \item \textbf{Tail entity:}
        If $e_{k'} \neq e_k$, then $\bm{e_{k'}^{F'}}$ is not changed if $e_{i'} = e_i$ and $\tau_{l'} = \tau_l$, and decremented by at most $2C$ at dimension $(j, k, l)$ otherwise. Hence, the changes to both $\bm{r_{j}^{t}}$ and $\bm{r_{x}^{t}}, x \neq j$ in Steps~4 \& 5 are sufficient to maintain $\bm{e_{k'}} \in \bm{r^{t}}$. 
        Conversely, if $e_{k'} = e_k$, then $\bm{e_t^{F'}}$ is unchanged when $e_{i'} \neq e_i$ and $\tau_{l'} = \tau_l$, and thus $\bm{e_{k'}} \in \bm{r^{t}}$ still holds. Otherwise, when $e_{i'} = e_i$ and $\tau_{l'} = \tau_l$, $\bm{e_{k'}^{F'}}$ is incremented by $C$, which, for $r = r_j$, makes $F'$ false, as required, and for $r \neq r_j$, still keeps $\bm{e_{k'}} \in \bm{r^{t}}$, as all other tail boxes are grown upwards by C.
        If $e_{k'} = e_k$, $e_{i'} = e_i$, and $\tau_{l'} \neq \tau_l$, then Step 3 ensures that $\bm{e_{k'}^F}$ remains unchanged, preserving the truth value of the fact.
    \end{enumerate}
    Therefore, any true fact $F' \neq \temptriple{e_i}{r_j}{e_k}{\tau_l}$ does still hold true after the inductive step.
    
    \item [] Case 2. \textbf{The fact $F'$ is false}: We show that this fact remains false after the inductive step by, again, analysing both the head entity $\bm{e_{i'}^{F'}}$ and tail entity $\bm{e_{k'}^{F'}}$.
    \begin{enumerate}[label=(\alph*)]
        \item \textbf{Head entity:} By construction, all false facts $\temptriple{e_{i'}}{r_{j'}}{e_{k'}}{\tau_{l'}}$ are made false by pushing their tail entity outside its target box, thus the following inequality is satisfied:
        \[
        \bm{e_{k'}^{F'}} (j', k', l') > \bm{u_{j'}^{t}} (j', k', l')
        \]
        Any changes to the head entity $\bm{e_{i'}^{F'}}$ do not affect this.
        \item \textbf{Tail entity:}
        If $e_{k'} \neq e_k$, then $F'$ verifies $\bm{e^{F'}}(j', k', l') > \bm{u_{j'}^{t}}(j', k', l')$, where $k' \neq k$. This inequality continues to hold regardless of the changes to $\bm{e_{k'}^{F'}}(j, k, l)$. Otherwise, if $e_{k'} = e_k$, and $r_{j'} = r_j$, then $e_{i'} \neq e_i$, as $F'$ is initially false, and $\temptriple{e_{i}}{r_{j}}{e_{k}}{\tau_{l}}$ is initially true in this induction step.  Furthermore, since $e_{i'} \neq e_i$, $\bm{e_{k'}^{F'}}$ is unchanged, which maintains the falsehood inequality. Finally, if $r_{j'} \neq r_j$, then the falsehood inequality for $F'$ holds at a dimension different than $(j, k, l)$. Therefore, none of the changes in the induction step affect this inequality.
    \end{enumerate}
    Hence, any false fact $F' \neq \temptriple{e_i}{r_j}{e_k}{\tau_l}$ is still false after the inductive step.
\end{itemize}

Therefore, it can be concluded that BoxTE is able to make any true fact false with $d=\left| \mathbf{R} \right| \left| \mathbf{T} \right| \left| \mathbf{E} \right|$.
It follows that BoxTE can represent any temporal knowledge graph $
\mathcal{G}$, and is fully expressive.
\qed

\subsection{Proof of \Cref{lem:REE}}
We modify the indexing scheme from above: For a vector $\bm{v} \in \mathbb{R}^d$ we write $\bm{v}(i, j, k)$ to refer to the index $i \left| \mathbf{E} \right| \left| \mathbf{R} \right|   + j \left| \mathbf{E} \right| + k$.
Intuitively, we assign one dimension in the embedding space to each possible non-temporal fact $\triple{e_i}{r_j}{e_k}$. Additionally, without loss of generality, we assume $k=1$ and $\alpha^{r_1}= ... = \alpha^{r_{\left| \mathcal{R} \right|}} = 1$, as in the proof for Lemma \ref{lem:RTE}.

Again, we show this property by induction. We start from a configuration that makes facts true if they hold in all time stamps, and makes any other fact false. That is,  we consider the static graph $\mathcal{G}' = \{\triple{h}{r}{t} \mid \forall \tau \in \mathbf{T} : \temptriple{h}{r}{t}{\tau} \in G \}$, where $\mathcal{G}$ is the original temporal knowledge graph.
Then, we use time bumps to make facts true without affecting the correctness/falsehood of other facts.

\paragraph{Base case:}
For a TKG $\mathcal{G}$ over $\mathbf{R}$, $\mathbf{E}$, $\mathbf{T}$ we consider the static graph $\mathcal{G}' = \{\triple{h}{r}{t} \mid \forall \tau \in \mathbf{T} : \temptriple{h}{r}{t}{\tau} \in G \}$.
Now we construct a BoxE configuration that captures the non-temporal $G'$, according to the BoxE full expressiveness proof \cite{BoxE-NeurIPS20}. However, for this construction, we use $d=\left| \mathbf{R} \right| \left| \mathbf{E} \right| \left| \mathbf{E} \right|$, and index each fact by the relation and both entities, resulting in indices of the form $(i, j, k)$. This is trivially possible, as this construction is analogous to the original BoxE proof, but introduces added dimensions and enables added indexing, which are necessary for proving this Lemma.

The resulting construction with $d=\left| \mathbf{R} \right| \left| \mathbf{E} \right| \left| \mathbf{E} \right|$, by design, is such that all false facts $F_f = \triple{e_i}{e_r}{e_k} \in \mathcal{G}'$ are made false by pushing the respective tail entity out of the tail box in dimension, i.e., $(i, j, k)$: $\bm{e^{F_f}_k}(i, j, k) > \bm{u_r^{t}}(i, j, k)$.

Observe that the BoxE construction results in a configuration where every false fact is false \emph{exclusively} at dimension $(i, j, k)$ (or $(i, k)$ in the original proof). Indeed, every induction step operates at index $(i, j, k)$, and preserves point membership at this dimension for both head and tail for all facts, except for the tail representation of the target fact. The induction step also does not affect any other dimensions. Thus, this construction does not cause additional falsehood beyond its target. Hence, to make a false fact $\triple{e_i}{e_r}{e_k}$ true, it is necessary and sufficient to adjust the representation at dimension $(i, j, k)$.

Given this configuration, we now introduce time bumps $\bm{\tau} = \bm{0}$ for all time stamps $\tau \in \mathbf{T}$ to obtain a BoxTE configuration in which a fact is true if and only if it is true for all time stamps in $\mathcal{G}$.

\paragraph{Induction step:}
In this step we consider a false fact $F = \temptriple{e_i}{e_r}{e_k}{\tau_l}$ and make it true without affecting the truth value of any other fact. We achieve this as follows:
\begin{itemize}
    \item[] \textbf{Step 1.} Define $C$ such that $\bm{e_k}(i, j, k) - \bm{u^{t}}(i, j, k)  < C < \bm{e_k}(i, j, k) - \bm{l^{t}}(i, j, k)$. 
    Decrement time bump $\bm{\tau_{l}}$ by $C$ in dimension $(i, j, k)$:
    \[
    \bm{\tau_{l}}(i, j, k) := \bm{\tau_{l}}(i, j, k) - C
    \]
    \item[] \textbf{Step 2.} For all relations $r_x \in \mathcal{R}$, grow both boxes downwards by $C$
    in dimension $(i, j, k)$:
    \[
    \bm{l_x^{h}}(i, j, k) := \bm{l_x^{h}}(i, j, k) - C
    \]
    \[
    \bm{l_x^{t}}(i, j, k) := \bm{l_x^{t}}(i, j, k) - C
    \]
\end{itemize}

Observe that Step $1$ pushes the tail entity of the false fact $F = \temptriple{e_i}{e_r}{e_k}{\tau_l}$ into its respective relation box:
\[\bm{e_k^{(e_i, r_j, e_k)}}(i, j, k) + \bm{\tau_l}(i, j, k) < \bm{u_j^{t}}(i, j, k)\]
\[\bm{e_k^{(e_i, r_j, e_k)}}(i, j, k) + \bm{\tau_l}(i, j, k) > \bm{l_j^{t}}(i, j, k) ,\]
therefore: 
\[\bm{e_k^{\temptriple{e_i}{e_r}{e_k}{\tau_l}}} \in \bm{r_j^{t}}\]
The change of time bump also affects the head entity, but its correctness is preserved by the box growing in Step 2. Hence, the target fact is made true, as required. We now show that the induction step does not affect the truth value of any other fact $F = \temptriple{e_{i'}}{e_{r'}}{e_{k'}}{\tau_{l'}}$:

\begin{itemize}
    \item [] Case 1. \textbf{The fact $F'$ is true}: We show that this fact remains true after the induction step by analysing both the head entity $\bm{e_{i'}^{F'}}$ and tail entity $\bm{e_{k'}^{F'}}$.
    \begin{enumerate}[label=(\alph*)]
        \item \textbf{Head entity:}
        Observe that in the induction step, the position of $\bm{e_{i'}^{F'}}$ changes by at most $C$ in dimension $(i, j, k)$, and only in the downward direction (when $l'=l)$.
        Therefore, growing the head box of every relation by $C$ downwards in the same dimension in Step 2 ensures that $\bm{e_{i'}^{F'}} \in \bm{r_{j'}^{h}}$ continues to hold.
        \item \textbf{Tail entity:}
        An analogous reasoning as for head entities applies to $\bm{e_{k'}}$ and $\bm{r_{j'}^{t}}$.
        
    \end{enumerate}
    Therefore, any true fact $F'$ remains true after the induction step.
    
    \item [] Case 2. \textbf{The fact $F'$ is false}: We also show that this fact remains false after the induction step by analysing the head entity $\bm{e_{i'}^{F'}}$ and tail entity $\bm{e_{k'}^{F'}}$.
    \begin{enumerate}[label=(\alph*)]
        \item \textbf{Head entity:}
        By construction, all false facts are false in their tail box, satisfying the inequality:
        \[
        \bm{e_{k'}^{F'}} (i', j', k') > \bm{u_{j'}^{t}} (i', j', k')
        \]
        Any changes to the head entity $\bm{e_{i'}^{F'}}$ do not affect this.
        \item \textbf{Tail entity:}
        Observe that Step 1 bumps all head boxes in the downward direction in dimension $(i, j, k)$.
        If $\temptriple{e_{i'}}{e_{r'}}{e_{k'}}{\tau_{l'}} = \temptriple{e_{i'}}{e_{r'}}{e_{k'}}{\tau_{l'}}$, i.e., $F'$ is the target fact, then $\bm{e_{k'}}$ gets bumped into $\bm{r_{j'}^{t}}$ as described above, making the fact true, as required.
        Otherwise, if $(i', j', k') \neq (i, j, k)$, i.e., different static indices, then, by construction, $\bm{e_{k'}}(i', j', k') > \bm{u_{j'}^{t}}(i', j', k')$, and this is not affected by any changes in dimension $(i, j, k)$. Finally, if $i=i', j=j', k=k'$, and $l \neq l'$ (same static components, different time stamp), then the representation of $\bm{e_k}$ is not affected by Step 1, and the box growth does not affect the upper bound, keeping the inequality at dimension $(i, j, k)$ false for this distinct timestamp.
        
    \end{enumerate}
    Hence, any false fact $F' \neq \temptriple{e_{i}}{e_{r}}{e_{k}}{\tau_{l}}$ is still false after the induction step.
\end{itemize}

Therefore, BoxTE is fully expressive with dimensionality $d= \left| \mathbf{R} \right| \left| \mathbf{E} \right| ^2$.
\qed

\section{Further Experimental Details}
\subsection{Hyper-parameter tuning}
In our main experiments, we trained BoxTE on ICEWS14, ICEWS5-15, and GDELT. For each hyper-parameter configuration, we trained models for 1200 epochs on ICEWS14 and ICEWS5-15 and 500 epochs on GDELT, and performed validation every 100 epochs, with peak MRR used as the validation metric.
For all experiments, we bound the embedding space to the hypercube $[-1, 1]^d$ by applying the hyperbolic tangent function $tanh$ element-wise to all final embedding representations.
Experiments were conducted on a single V100 GPU, on a compute node with 64 GB of RAM. In our tuning, the hyper-parameters considered include dimensionality, batch size, number of negative samples, and learning rate. However, we additionally experimented with the temporal smoothness regularizer from TNTComplEx, and tuned its weight hyper-parameter $\lambda$, and also considered factorizations of time embeddings $\mathbf{K^\tau}$. We explain this temporal smoothness regularizer and factorization next.

\begin{table*}[ht!]
    \centering
    \begin{tabular}{lccccccc}
        \toprule
        \emph{Standard setup} & \makecell{Embedding \\ Dimension} & \makecell{Learning \\ Rate} & \makecell{Negative \\ Samples} & \makecell{Batch \\ Size} & $k$ & $\lambda$ & \makecell{Factorisation \\ Basis $b$}\\
        \midrule
        ICEWS14           & 1000 & 0.001 & 75 & 256 & 2 & 0.1 & N/A \\
        ICEWS5-15         & 1500 & 0.001 & 75 & 512 & 5 & 0.1 & N/A \\
        GDELT      & 1500 & 0.001 & 75 & 256 & 5 & 0 & N/A \\
        \midrule
        \emph{Bounded-parameter setup} & & & & & & & \\
        \midrule
        ICEWS14        & 154 & 0.001 & 75 & 256 & 2 & 1 & N/A \\
        ICEWS5-15      & 104 & 0.001 & 75 & 256 & 3 & 0 & N/A \\
        GDELT     & 124 & 0.001 & 75 & 256 & 1 & 0.1 & 80 \\
        \bottomrule
    \end{tabular}
    \caption{Best hyper-parameter configuration of BoxTE on ICEWS14, ICEWS5-15, and GDELT.}
    \label{tab:hyperparams}
\end{table*}

\paragraph{Temporal smoothness regularizer.} TNTComplEx \cite{Lacroix2020Tensor} propose a temporal smoothness regularizer, which penalizes dissimilarity between temporal representations for consecutive time stamps, i.e., $\tau_i$ and $\tau_{i+1}$. Within BoxTE, we apply this regularizer to time embeddings $\mathbf{K}^\tau$ This regularizer is formally defined as:
\begin{equation*} \label{eq:temp-reg}
    \Lambda = \frac{1}{\left| \mathbf{T} \right| - 1} \sum_{i=1}^{\left| \mathbf{T} \right| - 1} \sum_{j=1}^k \left| \left| \mathbf{K^{\tau_{i+1}}}[j, :] - \mathbf{K}^{\tau_i}[j, :] \right| \right|_4 ^4,
\end{equation*}
where $[j, :]$ denotes the $j^\text{th}$ row vector of a matrix. This regularization term is then multiplied by a weight $\lambda$ and summed with model loss. Therefore, we additionally tune $\lambda$ in our experiments, and test values in the set $\{10^{-4}, 10^{-3}, 10^{-2}, 10^{-1}\}$.

\paragraph{Factorization.} In the main paper, we define an embedding matrix $\mathbf{K}^\tau$ for every time stamp $\tau \in \mathbf{T}$. When viewed holistically, however, these matrix stack up, and constitute an overall time embedding tensor $\mathbf{K} \in \mathbb{R}^{k \times |\mathbf{T}| \times d}$. Thus, to reduce parametrization, and enforce sharing across time stamps, we factorize $\mathbf{K}$ as the product of a left tensor $\mathbf{K}_L$ and a right tensor $\mathbf{K}_R$. More formally:
\begin{equation*}
    \mathbf{K} = \mathbf{K}_L \times_B \mathbf{K}_R,
\end{equation*}
where $\mathbf{K}_L \in \mathbb{R}^{k \times |\mathbf{T}|\times b}$, $\mathbf{K}_R \in \mathbb{R}^{k \times b \times d}$, $b\in \mathbb{N}^+$ is a tunable hyper-parameter, and $\times_B$ denotes batch matrix multiplication, with the first order as the batch dimension. In other words, we consider every one of the $k$ representations of all time stamps in $\mathbf{T}$, and factorize the resulting $|\mathbf{T}| \times d$ matrix as the product of a $|\mathbf{T}| \times b$ and a $b \times d$ matrix. 

To achieve parameter reduction, we experiment with values of $b$ from the set $\{20, 40, 80, 200, 400\}$ for all datasets.

In addition to these hyper-parameters, we also experimented with both cross-entropy loss and negative sampling loss \cite{RotatE-ICLR19}, and found the former to achieve slightly better results across all experiments. Furthermore, we tuned $d$ from the set $\{250, 500, 1000, 1500\}$ for the standard temporal knowledge graph completion experiments, and set $d$ to the correct parameter-bounding value for the bounded-parameter experiments. Finally, we tuned $k$ within the set $\{2, 3, 5\}$ in the standard TKGC setup ($\{1, 2, 3, 5\}$ in the bounded-parameter setup), the learning rate in the set $\{10^{-4}, 10^{-3}\}$, the number of negative samples per positive fact from the set $\{50, 75, 100, 150\}$, and batch size from the set $\{128, 256, 512, 1024\}$. 

The best hyper-parameter configurations across all datasets in our experiments are reported in \Cref{tab:hyperparams}.

\subsection{Model parameter counts}
\begin{table}[t!]
    \centering
    \begin{tabular}{lc}
        \toprule
        Model & Number of Parameters\\
        \midrule
        DE-SimplE       & $2d((3\gamma + (1 - \gamma)) \left| \mathbf{E} \right| + \left| \mathbf{R} \right|)$ \\
        TComplEx        & $2d(\left| \mathbf{E} \right| + \left| \mathbf{T} \right| + 2 \left| \mathbf{R} \right|)$ \\
        TNTComplEx      & $2d(\left| \mathbf{E} \right| + \left| \mathbf{T} \right| + 4 \left| \mathbf{R} \right|)$ \\
        \midrule
        BoxTE  & $d(2\left| \mathbf{E} \right| + k \left| \mathbf{T} \right| + 2 \left| \mathbf{R} \right|) + k \left| \mathbf{R} \right|$ \\
        BoxTE(F)  & $d(2\left| \mathbf{E} \right| + k b + 2 \left| \mathbf{R} \right|) + k \left| \mathbf{R} \right| + b k \left| \mathbf{T} \right|$  \\
        \bottomrule
    \end{tabular}
    \caption{Model parameter count for BoxTE and competing models. BoxTE(F) denotes the factorized BoxTE model, and thus additionally depends on the hyper-parameter $b$. For DE-SimplE, $\gamma$ denotes the share of temporal embedding features.}
    \label{tab:model_params}
\end{table}
In this subsection, we provide equations for computing the parameter counts of BoxTE~and competing models which report results in the parameter-bounded setting, relative to embedding dimensionality, $|\mathbf{E}|$, $|\mathbf{R}|$, and $|\mathbf{T}|$. For BoxTE, we provide two equations, corresponding to the non-factorised and factorized model configurations, respectively. The latter therefore depends on $b$, the factorization hyper-parameter.

All parametrization equations are provided in \Cref{tab:model_params}.

\section{Additional Experiments}
\subsection{Learned scalars on a subset of YAGO}
In this experiment, we develop a dataset from a subset of the YAGO knowledge graph, train BoxTE on this dataset, and observe the learned scalar vectors for every relation to evaluate whether BoxTE successfully captures relational temporal dynamics. We choose YAGO as it contains several prominent relations with substantial temporal semantics, e.g., $\mathsf{playsFor}$, and with varying degrees of temporal variability. For instance, the relation $\mathsf{playsFor}$ is not temporally stable, whereas $\mathsf{isMarriedTo}$ is.
\begin{table}[h]
    \centering
    \begin{tabular}{lcc}
        \toprule
        Relation & $\overline{\alpha}$\\
        \midrule
        playsFor        &   3.15\\
        participatedIn  & 1.84\\
        isAffiliatedTo  & 1.42\\
        hasWonPrize & 1.27 \\
        worksAt & 1.05 \\
        wroteMusicFor & 0.86 \\
        graduatedFrom & 0.74 \\
        isMarriedTo & 0.61 \\
        owns & 0.60 \\
        created & 0.59 \\
        \bottomrule
    \end{tabular}
     \caption{Average relational scalars $\overline{\alpha}$ learned on ICEWS49k-temp.}
     \label{tab:scalars}
\end{table}
\paragraph{Dataset construction.} We construct the dataset YAGO49k-temp from the larger YAGO dataset. To do so we consider the subset of YAGO15k that contains only temporal facts. Further we \emph{unfold} pairs of compatible facts with the tokens $\mathsf{ossursSince}$ and $\mathsf{occursUntil}$: From the dataset entries $(\mathsf{h, r, t, occursSince, \tau_1})$ and $(\mathsf{h, r, t, occursUntil, \tau_2})$ we construct the set of facts $\{ r(h, t | \tau) : \tau_1 \leq \tau \leq \tau_2 \}$.
By doing this we obtain facts that hold true for given ranges of time, which is unlike the more short-lived facts in GDELT and especially ICEWS14 and ICEWS5-15.
The statistics of the resulting datasat are shown in \Cref{tab:yago49kt_stats}.

\paragraph{Results.} The average scalar values for different relations in this dataset are shown in \Cref{tab:scalars}. Here, we observe that the temporally unstable relation $\mathsf{playsFor}$ learns large temporal scalars with $\bar{\alpha}=3.15$. Indeed, $\mathsf{playsFor}$ connects football players with teams, and each player changes his team 5.45 times over time, on average. This indicates a large degree of temporal variability, which is reflected in the associated scalar mean. By contrast, the $\mathsf{isMarriedTo}$ relation, which is more stable, as every person is on average married to 1.05 distinct partners over time, learns much smaller temporal scalars, averaging around $\bar{\alpha}=0.61$.

\begin{figure*}[t!]
\centering
\begin{subfigure}{.33\textwidth}
  \centering
  \includegraphics[width=1\linewidth]{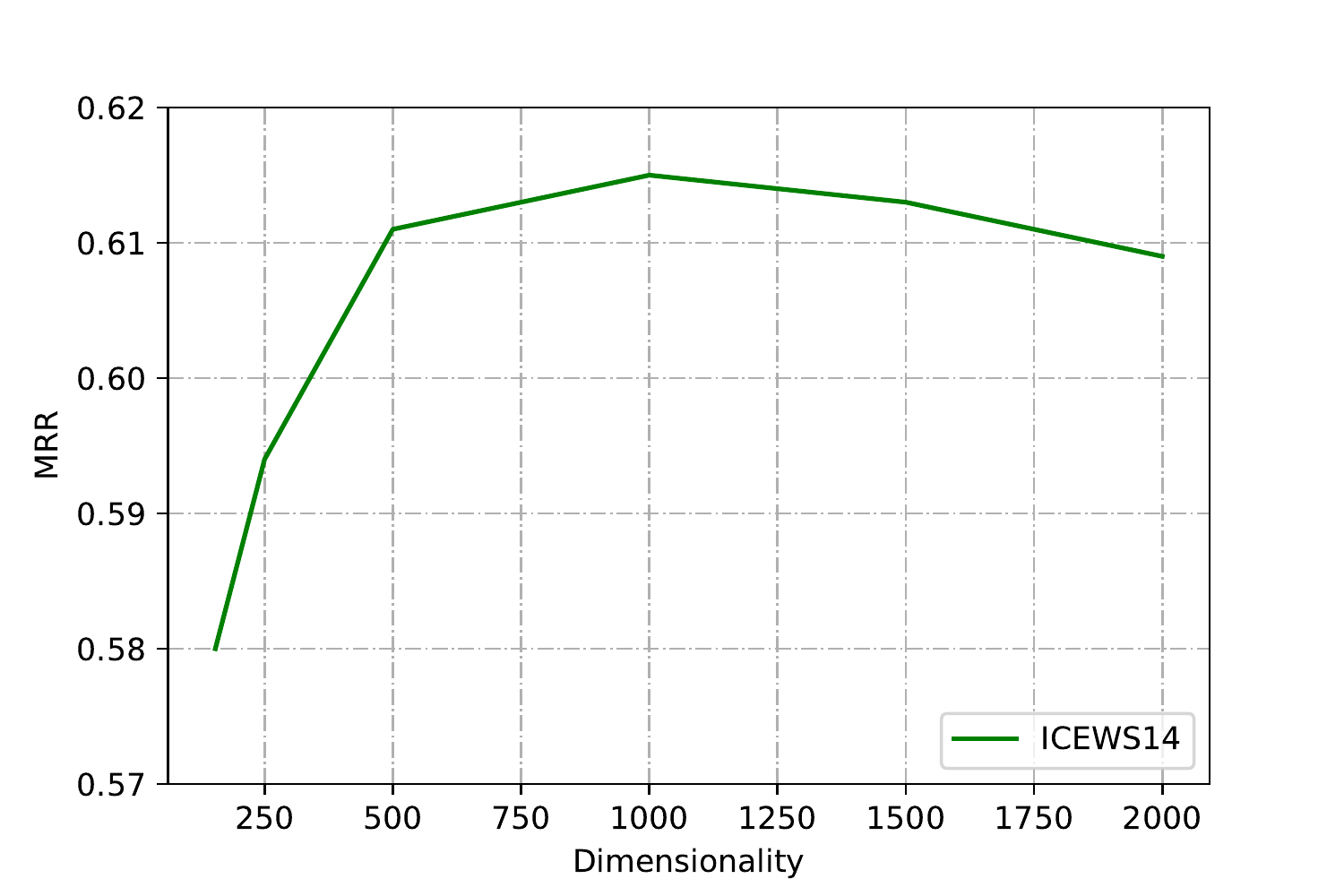}
  \caption{ICEWS14}
\end{subfigure}%
\begin{subfigure}{.33\textwidth}
  \centering
  \includegraphics[width=1\linewidth]{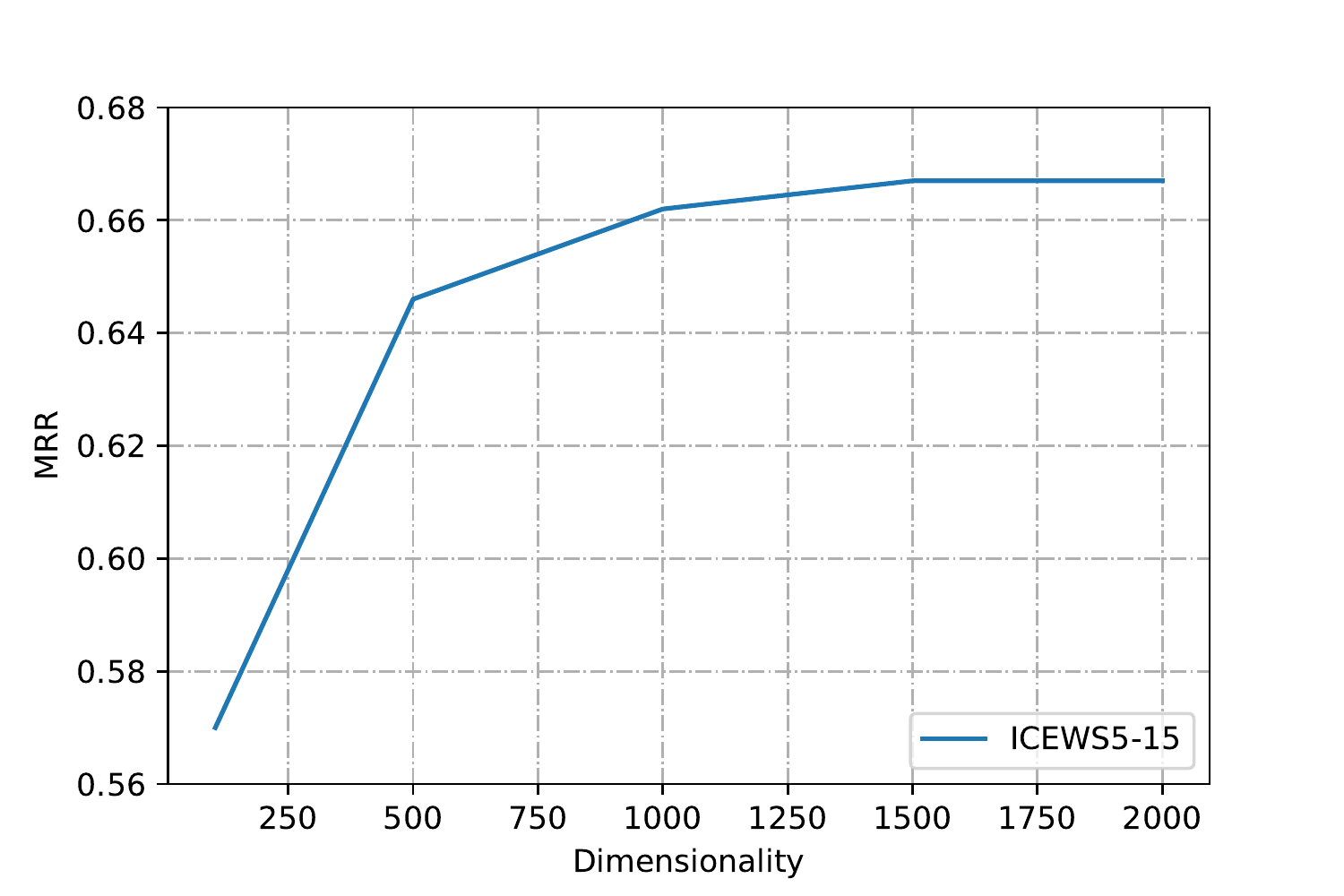}
  \caption{ICEWS15}
\end{subfigure}
\begin{subfigure}{.33\textwidth}
  \centering
  \includegraphics[width=1\linewidth]{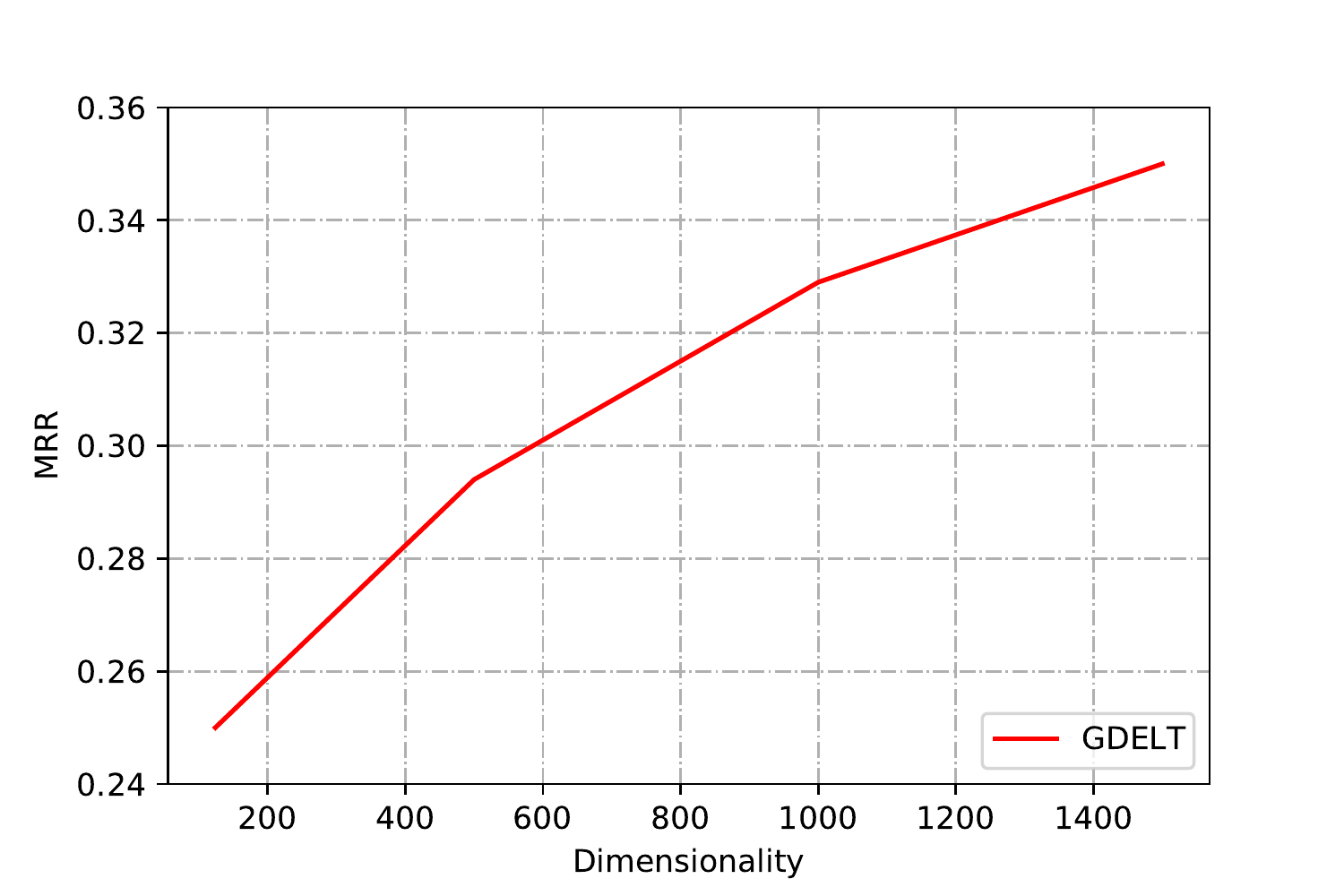}
  \caption{GDELT}
\end{subfigure}
\caption{Mean reciprocal rank (MRR) achieved with different embedding dimensionalities, on ICEWS14 (a), ICEWS5-15 (b) and GDELT (c) datasets.}
\label{fig:dim-v-mrr}
\end{figure*}

With the relations $\mathsf{participatedIn}$ and $\mathsf{isAffiliatedTo}$, a head entity can  simultaneously connect with multiple distinct tail entities, and thus the lack of mutual exclusion (as is the case with $\mathsf{playsFor}$) makes these relations less temporally variable. Nonetheless, these relations do exhibit significant variability over time. Accordingly, the model learns intermediate average scalar values for both relations, namely $\bar{\alpha}=1.84$ and $\bar{\alpha}=1.42$ respectively.

Additionally, the dataset contains several relations whose facts are true at single distinct moments in time, rather than over temporal ranges, such as $\mathsf{graduatedFrom}$, $\mathsf{created}$, $\mathsf{wroteMusicFor}$, and $\mathsf{hasWonPrize}$. As facts corresponding to these relations only hold true at one time stamp, and are false at any other time stamp, these relations learn a scalar configuration reflecting this stability, obtaining average scalar values of $\bar{\alpha}=0.74$, $\bar{\alpha}=0.59$, $\bar{\alpha}=0.86$ and $\bar{\alpha}=1.27$, respectively.
Hence, temporal scalars accurately reflect the temporal variability of a given relation in this experiment, with stable relations learning small scalars, and highly unstable relations learning larger scalars.

 \begin{table}[t]
    \centering
    \begin{tabular}{ll}
        \toprule
                           & YAGO49k-temp   \\
        \midrule
        $\left| \mathbf{E} \right|$ & 4139 \\
        $\left| \mathbf{R} \right|$ & 10\\
        $\left| \mathbf{T} \right|$ & 506 \\
        $\left| \mathcal{G}_\text{train} \right|$ & 49509\\
        $\left| \mathcal{G}_\text{valid} \right|$ &  6174\\
        $\left| \mathcal{G}_\text{test} \right|$ & 5964\\
        Timespan           & 1417 years     \\
        Granularity        & Yearly      \\
        \bottomrule
    \end{tabular}
    \caption{YAGO49k-temp dataset statistics.}
\label{tab:yago49kt_stats}
\end{table}

\subsection{Robustness analysis}

To investigate the robustness of our model relative to embedding dimensionality $d$, we train BoxTE on ICEWS14, ICEWS5-15 and GDELT while varying $d$ and report the best result at each value.

As figure \ref{fig:dim-v-mrr} shows, BoxTE achieves state of the art results on ICEWS14 in the constrained parameter setting with $d=154$, and comes within 1\% of its best possible performance with $d=500$.
Peak performance is achieved with $d=1000$, while higher dimensionalities show a slight decrease in performance.
On ICEWS5-15 a similar picture emerges, where $d=500$ suffices to come within 0.02 of the best possible result, in terms of MRR. Here we observe no overfitting, even with $d=2000$.
Overall, we conclude that BoxTE is robust with regard to the chosen embedding dimensionality, with strong performance on all settings between 500 and 2000 dimensions.

On the much larger GDELT dataset we find that MRR increases across the entire range  of computationally feasible dimensionality values. This suggests that BoxTE, in our testing, may not yet have reached its best possible performance on this dataset.
This further highlights the complexity and density of information in GDELT dataset.

\subsection{Ablation studies}
In order to contextualise the results obtained by BoxTE, we perform four ablation studies.
First, we investigate the effect of the temporal regularizer on our and competing models.
Then, we use competing approaches to make static BoxE temporal and compare with their results.
Thirdly, we train BoxTE without relation-specific temporal scalars, and lastly, we compare results obtained by training with different loss functions.

\paragraph{Temporal regulariser.}
\begin{figure}[t]
\centering
\begin{subfigure}{.4\textwidth}
  \centering
  \includegraphics[width=1\linewidth]{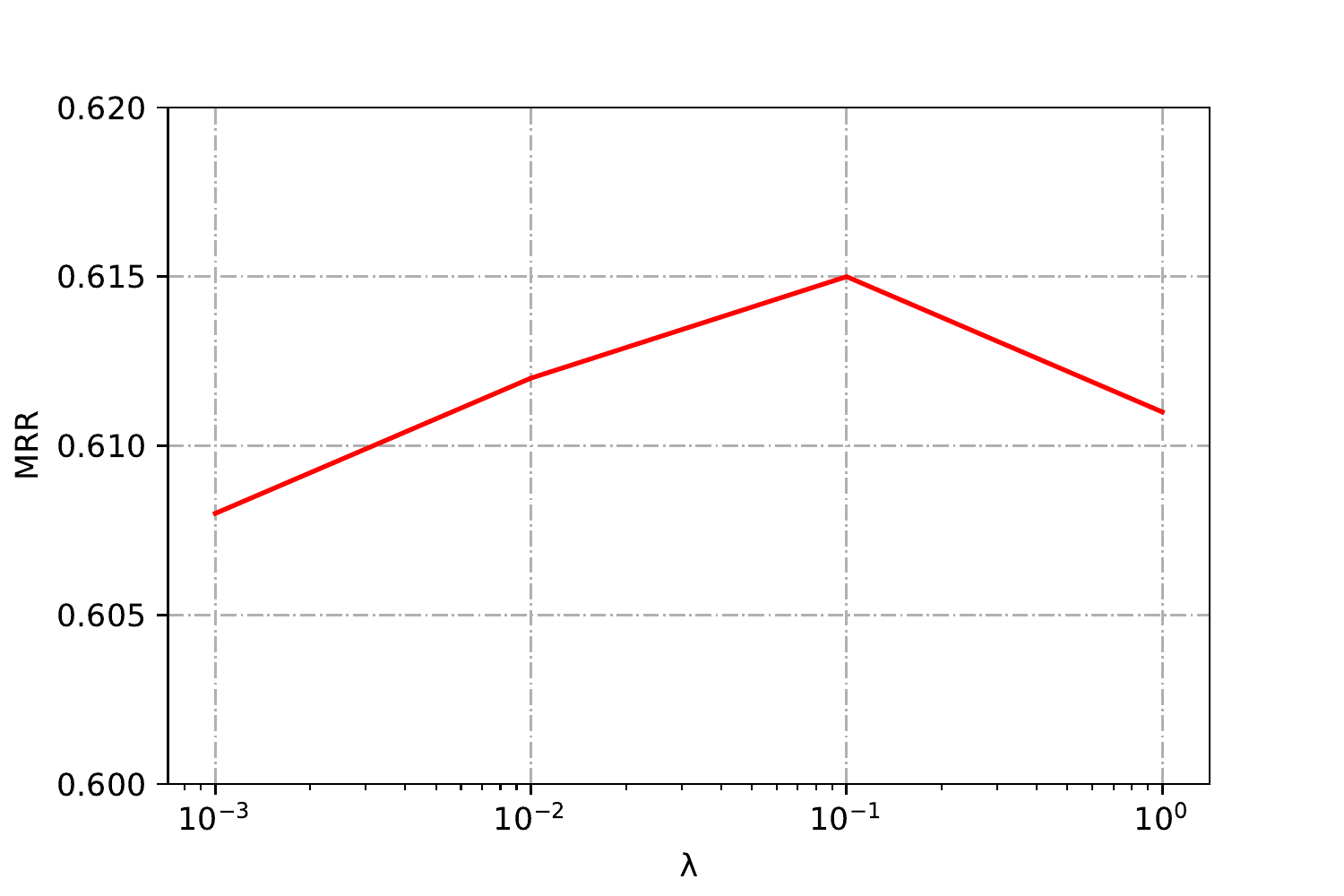}
  \label{fig:timereg-v-mrr}
\end{subfigure}%
\\
\begin{subfigure}{.4\textwidth}
  \centering
  \includegraphics[width=1\linewidth]{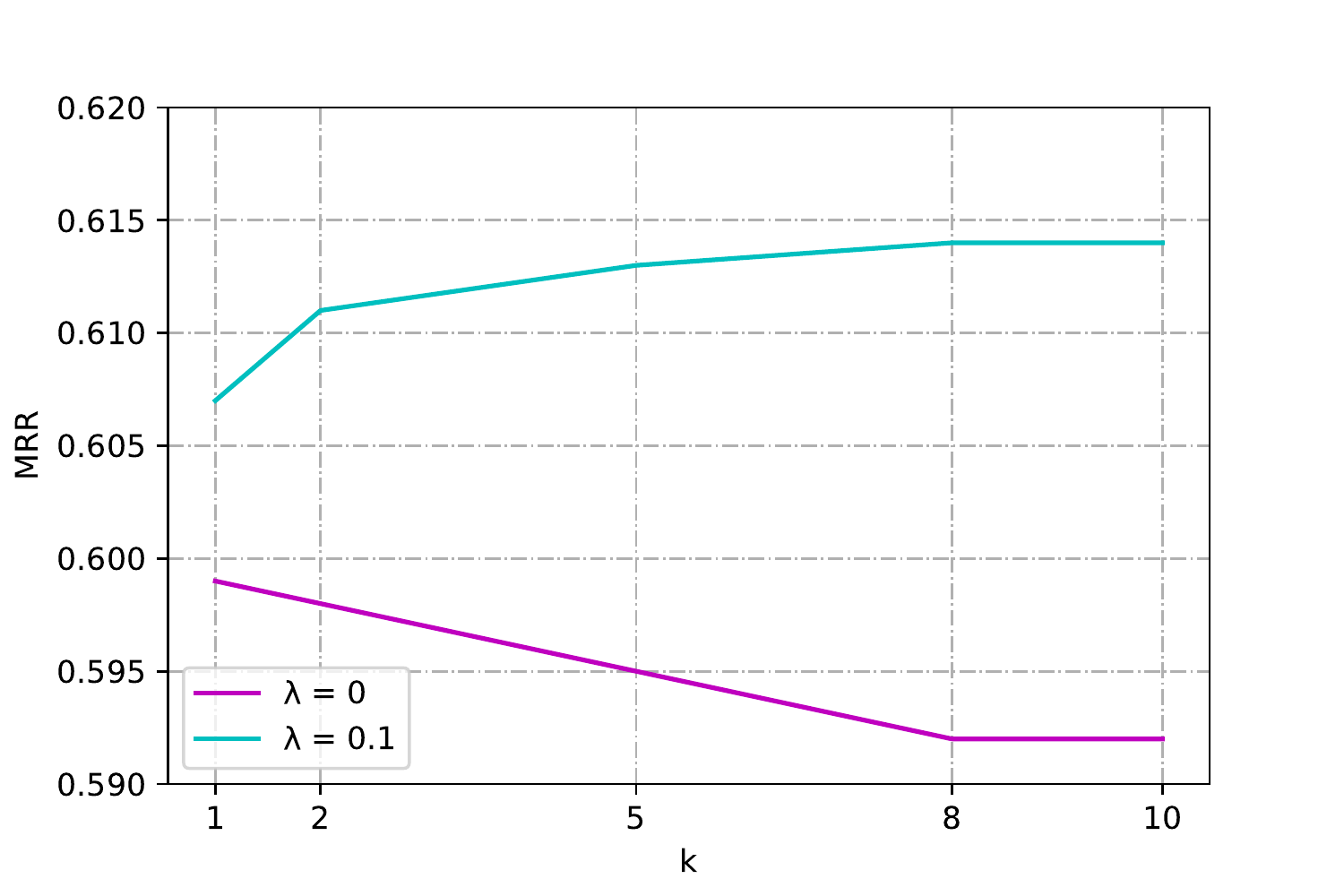}
\end{subfigure}
\caption{Influence of temporal regularizer on MRR on ICEWS14 dataset. (a) BoxTE maintains strong performance even without regularisation with $k=2$, $d=1000$. (b) The temporal regularizer prevents overfitting with large $k$, $d=500$.}
\label{fig:k-v-mrr-timereg}
\end{figure}
To better understand the impact of regularisation on the performance of TKGC models, we additionally conduct experiments on both ICEWS datasets where we omit this regularizer (\ref{eq:temp-reg}), and otherwise choose our best performing configuration.

There, we observe that competing models suffer considerably, whereas BoxTE maintains very strong performance, and in fact becomes the best-performing model. Indeed, BoxTE achieves an MRR of 0.608 on ICEWS14, while the best ICEWS14 models ChronoR \cite{ChronoR} and TeLM \cite{xu-etal-2021-temporal} each report a value of less than 0.6.
On ICEWS5-15, BoxTE achieves an MRR of 0.657 and thus surpasses TNTComplEx \cite{Lacroix2020Tensor}, which reports an MRR of less than 0.65. This suggests that these models are more dependent on the regularisation prior to learn temporal patterns, whereas BoxTE is more robust, and can partly recover this information even when it is not provided. This further highlights the importance of BoxTE semantics, which provide a more holistic means of capturing temporal patterns. This is especially relevant beyond our evaluation setting, where such regularisation terms could not be known, or could not be compatible with the given dataset. In fact, on GDELT, temporal regularisation is detrimental to performance, which suggests that it this regularisation is incompatible with the input data. Indeed, GDELT has large temporal variability, where facts can hold for extended periods of time, but also hold rarely, and thus a general smoothness assumption conflicts with this information and yields worse performance, as expected.

\paragraph{Model variants.}
\begin{table}[t!]
    \centering
    \begin{tabular}{lc}
        \toprule
        Model variant & MRR\\
        \midrule
        TTransE         & .255 \\
        TBoxE           & .455 \\
        \midrule
        DE-SimplE       & .526 \\
        DE-BoxE         & .476 \\
        \midrule
        BoxTE    & .615 \\
        \bottomrule
    \end{tabular}
    \caption{Performance (MRR) of model variants BoxTE, TBoxE, DE-BoxE on ICEWS14.}
    \label{tab:results_model_variants}
\end{table}
We instantiate two BoxE variants that directly implement competing time representation approaches in BoxE space.

\emph{TBoxE} implements the approach introduced by TTransE \cite{TTransE}.
In TTransE, each time stamp is represented as an additional relation embedding that acts on entity embeddings.
Analogously, we implement a model where each time stamp $\tau$ is represented by a head box $\bm{\tau^{h}}$ and a tail box $\bm{\tau^{t}}$.
TBoxE then scores a fact as follows:
\[
score(\temptriple{h}{r}{t}{\tau}) = \sum_{x \in \{h, t\}} \sum_{b \in \{r, \tau\}} \delta (\bm{e_x^{\temptriple{h}{r}{t}{\tau}}}, \bm{b^{(x)}})
\]
Essentially, each time stamp gets treated like an additional relation, and the scoring function sums the distance over all four boxes in a fact.

\emph{DE-BoxE} is an application of the approach to time representation introduced by DE-SimplE \cite{DiachronicEmbedding}.
This model variant directly applies the diachronic embedding framework:
For each entity embedding vector, a proportion $\gamma$ of dimensions is being reserved for the representation of temporal variations.
These dimensions are trained as usual, but then passed through  an activation function ($sine$, $\sigma$) and projected through time by multiplying the time index associated with a fact.
We experiment with applying this approach to final entity embeddings, entity bumps, and entity bases.
We find that DE-BoxE performs best when diachronic embeddings are used on the entity bases, and report results accordingly.
Adopting this approach in BoxE space does not require any alterations to the scoring scheme.

As we show in Table \ref{tab:results_model_variants}, TBoxE and DE-BoxE underperform significantly when compared to BoxTE, thus indicating that the strong results obtained by BoxTE are a result of the novel time representation, and not merely of the expressive BoxE model that underlies it.
Furthermore, TBoxE outperforms TTransE, as is expected due to the stronger static model.
Interestingly, however, DE-BoxE is unable to improve on or even match the performance of DE-SimplE.
This can be explained by considering BoxE's semantics and distance function.
In BoxE, a fact is considered to be true if and only if both entity embeddings are inside their respective target boxes, in all dimensions.
In the DE-approach, the choice of activation function imposes a prior on how temporal features are expected to behave over time: A $sine$ activation function presupposes a periodic evolution of temporal features, while a sigmoid $\sigma$ assumes monotonic growth.
If these assumptions do not precisely agree with the data, it is unlikely that every temporal dimension can be configured in such a way that all entities are placed in their respective target boxes, on all relevant time stamps.
Thus, TBoxE struggles to fit the data adequately, and ultimately underperforms.

\paragraph{Temporal scalars.}
\begin{table}[t!]
    \centering
    \begin{tabular}{lc}
        \toprule
        Model & MRR\\
        \midrule
        $k=2$, temporal scaling           & .615\\
        $k=1$, temporal scaling          & .614\\
        no temporal scaling      & .571\\
        \bottomrule
    \end{tabular}
    \caption{Model performance with modular time bumps (default setting, $k>1$), scaling only ($k=1$), and relation-independent bump (disabled temporal scalars), with $d=1000$, on ICEWS14.}
    \label{tab:no_scaling}
\end{table}
We disable learnable time scalars and train BoxTE on ICEWS14, leaving all other parameters unchanged compared to our best performing configuration.
Disabling temporal scalars prohibits relations from modulating the magnitude and direction of a given time bump.
We additionally compare to a setup with $k=1$ and enabled scalars. We report out results in Table \ref{tab:no_scaling}.

We find that on ICEWS14, unlike ICEWS5-15, employing just a single temporal bump per snapshot ($k=1$) does not lead to a significant drop in performance.
However, additionally disabling the ability of relations to scale time bumps leads to a drop in MRR of $.043$.
This finding underlines the importance of allowing relations to modulate time bumps, in order to accurately capture how relations are affected differently by time.

\begin{figure}
    \centering
    \includegraphics[width=0.4\textwidth]{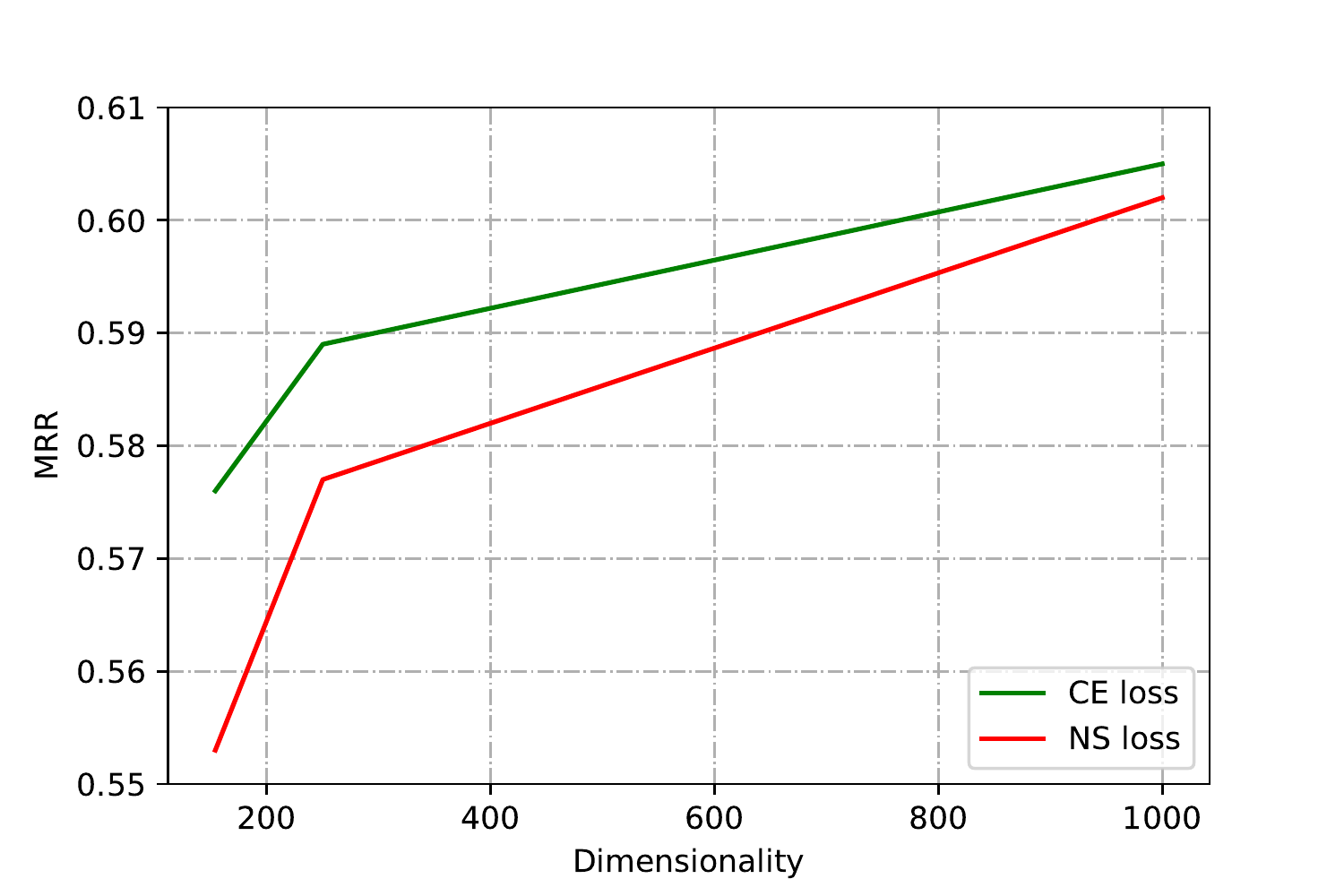}
    \caption{Mean reciprocal rank (MRR) achieved with cross entropy loss (CE) and adversarial negative sampling loss (NS) on ICEWS14, with $dim \in \{154, 250, 1000 \}$, $k=2$, $\lambda=0$.}
\label{fig:losses-icews14}
\end{figure}

\paragraph{Loss function.}

To compare between loss functions, we train BoxTE with cross-entropy loss and self-adversarial negative sampling loss \cite{RotatE-ICLR19} on the ICEWS14 dataset.
Results are shown in Figure \ref{fig:losses-icews14}.
As the figure shows, we find that cross entropy loss consistently outperforms negative sampling loss.
We also observe that this difference decreases as dimensionality grows: With 154 dimension the difference is 4.2\% or .023 points in MRR, with 1000 dimension it is just 0.5\% or .003 MRR points.
We also note that adversarial negative sampling loss requires tuning of two additional hyperparameters, the \emph{margin} $\gamma$ and the \emph{adversarial temperature} $\alpha$.
Overall we conclude that cross entropy loss is the better suited objective function for our specific experimental setting.  %

\end{document}